\documentclass{article} 
\usepackage{iclr2027_conference,times}
\usepackage{graphicx}

\usepackage{amsmath,amsfonts,bm}

\def\eqref#1{equation~\ref{#1}}

\def\1{\bm{1}}

\DeclareMathAlphabet{\mathsfit}{\encodingdefault}{\sfdefault}{m}{sl}
\SetMathAlphabet{\mathsfit}{bold}{\encodingdefault}{\sfdefault}{bx}{n}

\usepackage{hyperref}
\usepackage{booktabs}
\usepackage{url}

\title{A Low-Rank Space Captures Linear \\ Alignment between LLM and Human Neural Activity during Naturalistic Language \\ Listening}

\author{Zach Paris \& Emily Finn \\ 
Department of Psychological and Brain Sciences\\
Dartmouth College\\
Hanover, NH 03755, USA \\
\texttt{\{zachary.r.paris.gr, Emily.S.Finn\}@dartmouth.edu} \\
}

\iclrfinalcopy 

\begin{document}

\maketitle

\begin{abstract}
Large language models (LLMs) are increasingly used as computational models of human language processing, yet it remains unclear which components of their representations drive alignment with neural activity. Here, we use ridge reduced-rank regression to identify the brain-predictive subspace of LLM representations: the representational directions that support linear prediction of neural responses. Across three naturalistic language-listening datasets spanning intracranial electrophysiology and fMRI, and six model conditions spanning autoregressive, bidirectional masked, and randomly initialized language models, we find that full linear encoding performance can be recovered using only ~1–3\% of the original embedding dimensionality. These low-dimensional subspaces generalize across held-out participants and, to a lesser extent, across datasets, suggesting that neural predictivity is concentrated in a compact and partially shared representational space. We next compare these brain-predictive directions with the principal axes of variation in the embeddings. Principal component analysis requires substantially more dimensions to recover neural prediction, demonstrating that the directions most useful for predicting neural activity are distinct from those capturing the most overall variance within LLM representations themselves. Finally, characterizing the learned subspaces allows us to surface interpretable features that are overrepresented in neural activity above and beyond the LLM representations. Together, these results show that LLM–brain alignment is supported by a small, structured subset of model representations and provides a framework for better understanding exactly why these representations align. Code can be found at: \href{https://github.com/thefinnlab/low-rank-alignment}{https://github.com/thefinnlab/low-rank-alignment}

\end{abstract}

\section{Introduction}

Large language models (LLMs) are strong predictors of human neural activity during language processing and are increasingly used as computational models of language comprehension \citep{schrimpfNeuralArchitectureLanguage2021, goldsteinSharedComputationalPrinciples2022, mischlerContextualFeatureExtraction2024, zadaSharedModelbasedLinguistic2024}. Yet it remains unclear which components of their high-dimensional representations drive this alignment. Prior work has related neural predictivity to properties such as training objective, model scale, layer depth, intrinsic dimensionality, and representational generality \citep{antonelloScalingLawsLanguage2023, antonelloPredictiveCodingJust2024, antonelloEvidenceFMRISupports2024}. However, these model-level relationships do not identify the specific representational directions that support neural prediction, and apparent differences in alignment can also be influenced by confounds in the evaluation process \citep{hadidiSpuriousAlignmentLarge2026}.

Here, we take a complementary approach. Rather than asking which global properties of a model correlate with its neural predictivity, we ask which part of the model representation actually supports that prediction. Conceptually, we view LLM–brain alignment as arising from information shared between features represented by an LLM and measured neural responses (figure \ref{fig:methods}a). Under a linear encoding model, we operationalize this shared information as a brain-predictive subspace of the LLM representation: the set of representational directions that are useful for predicting neural activity. This framing turns a single encoding score into a representational object that can itself be studied.

Across three naturalistic language-listening datasets spanning intracranial electrophysiology and fMRI, and six different language models we use reduced rank regression (RRR) \citep{izenmanReducedrankRegressionMultivariate1975} to discover this brain-predictive subspace and find that full linear encoding performance can be recovered from only ~1--3\% of the original embedding dimensions. These brain-predictive subspaces generalize across held-out participants and, partially, across datasets.  We then compare each learned subspace against one learned by Principal Component Analysis (PCA) applied only to the LLM embeddings. We find that there is a mismatch between the directions of principal variance in the embeddings as learned from an unsupervised PCA and the directions most useful for neural prediction; suggesting differences in the representational geometry of the two systems. Finally, feature-decoding analyses reveal interpretable differences in the linguistic information emphasized by these subspaces across datasets and recording modalities. Together, these results show that LLM--brain alignment is supported by a compact, structured subset of model representations and provide a framework for identifying which components of those representations drive neural predictivity.

\section{Background and Motivation}
\subsection{Language Encoding Models}

Encoding models test whether a candidate representational space captures information present in neural responses. In language neuroscience, this approach has progressed from investigator-defined semantic features to contextual representations learned by neural language models, which predict activity across fMRI, MEG, and intracranial recordings \citep{huthNaturalSpeechReveals2016,jainIncorporatingContextLanguage2018,tonevaInterpretingImprovingNaturallanguage2019,schrimpfNeuralArchitectureLanguage2021,goldsteinSharedComputationalPrinciples2022,caucheteuxBrainsAlgorithmsPartially2022}.

LLM representations are effective encoding spaces because they capture many aspects of language simultaneously, but this also complicates their interpretation. A single layer may contain thousands of distributed features reflecting lexical, syntactic, semantic, contextual, and predictive information, any combination of which may support neural prediction. Thus, successful encoding establishes that neurally relevant information is linearly accessible from the representation but does not reveal which components carry that signal.

Prior work has related neural predictivity to properties including next-word prediction performance, model scale, layer depth, representational generality, intrinsic dimensionality, and particular linguistic or computational features \citep{schrimpfNeuralArchitectureLanguage2021,antonelloPredictiveCodingJust2024,antonelloEvidenceFMRISupports2024,kumarSharedFunctionalSpecialization2024,mischlerContextualFeatureExtraction2024}. These results constrain possible explanations of LLM--brain alignment but primarily characterize representations globally rather than identifying the specific directions responsible for prediction.

\subsection{Brain-Predictive Subspaces}

Prior work suggests that both language representations and their relationship to neural activity exhibit low-dimensional structure. \citet{antonelloLowdimensionalStructureSpace2021} identified low-dimensional organization in linguistic representations related to fMRI predictivity, while \citet{lewisLowRankTensorEncoding} used low-rank constraints to factorize LLM-to-MEG encoding models into interpretable semantic, temporal, and spatial components.

These approaches motivate a more direct question: how much of an individual LLM representation is actually required to recover its neural predictivity? Reduced-rank regression provides a supervised way to answer this question by identifying low-dimensional directions chosen specifically for their ability to predict neural responses. This distinguishes the brain-predictive subspace from unsupervised low-dimensional structure, such as the principal variance directions identified by PCA, and provides a representation whose dimensionality, generalization, and linguistic content can be studied directly.

\section{Methods}

\begin{figure}[t!]
\begin{center}
\includegraphics[width=\textwidth]{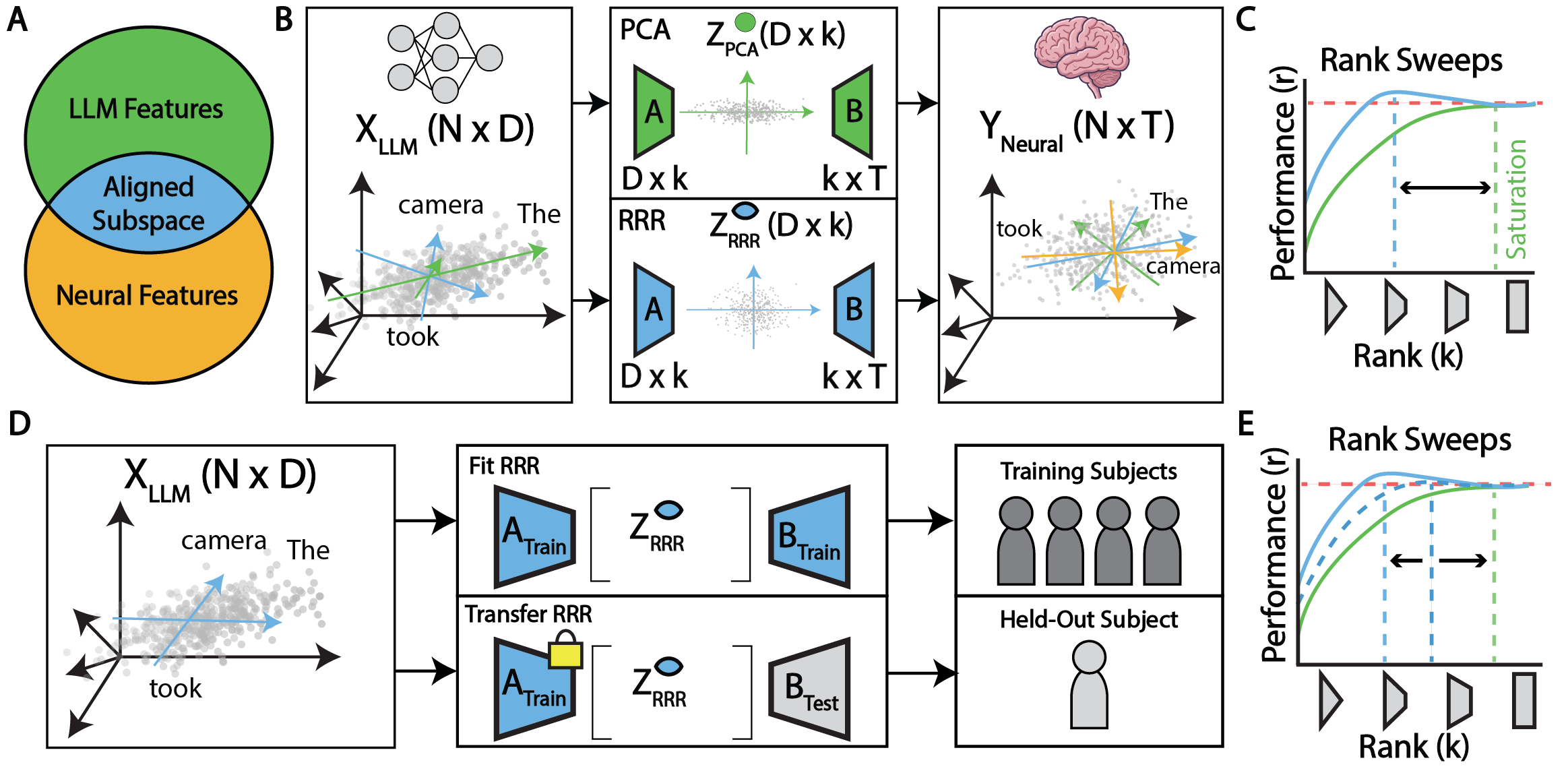}
\end{center}
\caption{\textbf{(a)} We view alignment as a measure of the linear features shared by the LLM embedding space and the neural response space. Our goal is to selectively extract the intersection of the two spaces. \textbf{(b)} Our pipeline. We learn two low-rank representations of the model embeddings and use them to predict neural data. The PCA space (green) prioritizes capturing the directions of principal variation in the LLM embeddings. The RRR space (blue) prioritizes the directions which best align between the two representational spaces teasing out the aligned subspace. \textbf{(c)} To identify the size of the aligned subspace we sweep the subspace rank, fit new models and re-evaluate. We find the point at which our RRR space performance saturates and compare it to the PCA point of saturation. \textbf{(d)} Leave-one-subject-out experiment. We  fit the RRR space on all but one subject and then transfer the projection matrix to a held-out subject and fit a new mapping to that subject's neural data. \textbf{(e)} We again sweep over ranks and compare the point of saturation on the held-out subject between the RRR space fit on all subjects, RRR space fit on all but the held-out subject, and the unsupervised PCA space.}
\label{fig:methods}
\end{figure}

\subsection{Datasets}

We analyzed three datasets of neural responses to naturalistic spoken language. The Podcast ECoG dataset contains high-gamma (70--200 Hz) recordings from nine participants and 1,330 intracranial electrodes collected during a 30-minute narrative podcast \citep{zadaPodcastECoGDataset2025}. Brain Treebank contains stereoelectroencephalography (sEEG) recordings from participants viewing movies \citep{wangBrainTreebankLargescale2024}; we analyzed \textit{Cars 2}, which contained the most heard words ($\sim$8,000) viewed by at least three participants. The Deep fMRI dataset contains responses to spoken stories collected across five recording sessions \citep{lebelNaturalLanguageFMRI2023}; we analyzed eight participants and evaluated encoding models on the repeated held-out test story.

\subsection{Preprocessing}

We used the released high-gamma responses for Podcast and preprocessed Brain Treebank recordings to extract 70--200 Hz high-gamma activity. Both intracranial datasets were aligned to word onset and sampled at 32 Hz. We used the released preprocessed fMRI responses. Full preprocessing details are provided in Appendix A.1.

\subsection{Language models}

We extracted word representations from four autoregressive models---GPT-2, GPT-2 XL, GPT-J 6B, and Llama 3 8B---and the bidirectional masked language model RoBERTa Large \citep{radford2019language,gpt-j,llama3modelcard,liuRoBERTaRobustlyOptimized2019}. We additionally analyzed a randomly initialized GPT-2 control as such controls have been studied in previous work as well as a non-contextual 300-dimensional GloVe embeddings \citep{schrimpfNeuralArchitectureLanguage2021,hosseiniArtificialNeuralNetwork2024,pennington2014glove}. For each contextual model, we sampled five layers at 0\%, 25\%, 50\%, 75\%, and 100\% depth. Additional embedding-generation details are provided in Appendix A.1.1.

\subsection{Selection of neural targets}

Because our goal was to estimate the rank of the subspace supporting general neural alignment, we used a broad target-selection procedure designed to retain reliable signal linking LLM embeddings to neural activity while excluding clearly noisy targets. We selected recording sites using held-out predictions from full-embedding ridge encoding models. Significance was assessed against null distributions generated by circularly shifting neural responses, with Benjamini--Hochberg correction across recording sites. For iEEG, selected electrodes contributed all sampled lags as RRR targets; for fMRI, selected voxels were used directly. Additional selection details and target counts are provided in Appendix A.5.

\subsection{Encoding models}

We first fit ridge encoding models using the full embedding space to establish a reference performance for each model and layer. For iEEG, word embeddings predicted electrode responses at sampled lags relative to word onset, with performance evaluated using five contiguous held-out word folds. For fMRI, embeddings were aligned to acquisition times and used to construct a four-delay finite impulse response design; models were evaluated on held-out stories. Ridge penalties were selected by cross-validation.

\begin{figure}[t]

\begin{center}
\includegraphics[width=\textwidth]{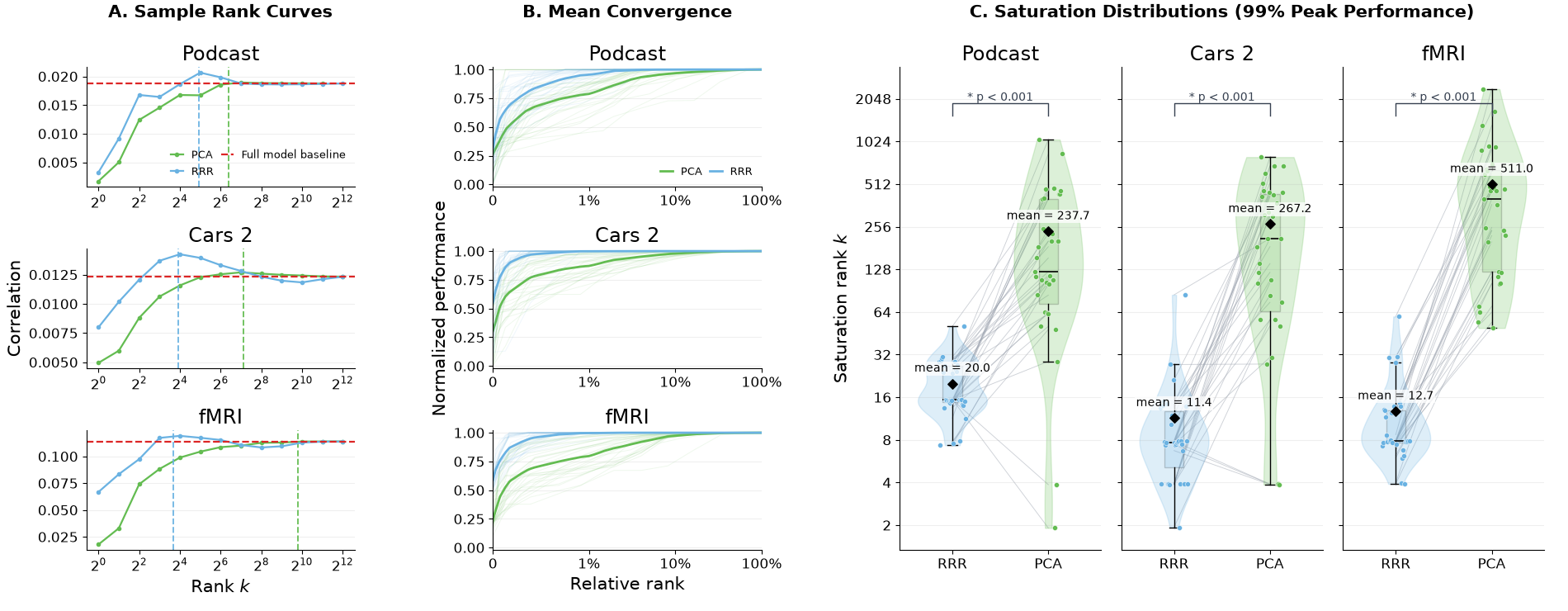}
\end{center}
\caption{\textbf{Low-rank RRR recovers neural predictivity more efficiently than PCA.}
\textbf{(a)} Example rank-performance curves for GPT-J-6b layer 21 across the three datasets. \textbf{(b)} Mean convergence across model-layers showing RRR converges quicker than PCA. Generated by normalizing by max performance and enforcing monotonicity before averaging. \textbf{(c)} Saturation ranks for matched model--layer combinations, defined as the minimum rank at which all informative target-performance quantiles reach 99\% of their maximum. RRR saturates at substantially lower rank than PCA across all three datasets. See Methods and Appendix for threshold and significance details.}
\label{fig:low}
\end{figure}

\subsection{Reduced-rank regression}

We used ridge reduced-rank regression (RRR) \citep{izenmanReducedrankRegressionMultivariate1975} to identify the optimal rank $k$ subspace of embeddings for neural prediction (Figure \ref{fig:methods}b). Within each training partition, we fit a multivariate ridge model from embeddings ($X$) to neural responses ($Y$), producing coefficients ($W$) and fitted responses ($\hat{Y}=XW$). We decomposed ($\hat{Y}$) and used its leading ($k$) right singular vectors ($V_k$) to construct an input factor ($A_k=WV_k$), whose columns span embedding directions associated with the leading components of predicted neural activity.

At each rank ($k$), we projected embeddings into an orthonormalized version of this subspace and fit a new ridge encoding model using the projected data. We swept ranks from one to the available embedding-dimensionality endpoint to determine how quickly neural prediction was recovered as the bottleneck increased (figure \ref{fig:methods}c). Performance was evaluated on held-out neural responses, with the full-embedding ridge model serving as the reference. For fMRI, we combined the delay-specific RRR factors into a single orthonormal basis in the original embedding coordinates; details are provided in  Appendix A.2.2.

We compared RRR with principal component analysis (PCA) of the training embeddings (figure \ref{fig:methods}b). At each rank, embeddings were projected onto the leading principal components and evaluated using the same ridge refitting and held-out data.

\subsection{Saturation rank}

To summarize how quickly predictive performance was recovered as rank increased (figure \ref{fig:methods}c), we defined a saturation rank using the distribution of performance across neural targets. At each rank, we computed the 10th, 25th, 50th, 75th, and 90th percentiles of held-out target performance and tracked each quantile across the rank sweep. Saturation rank was defined as the minimum rank at which all informative quantile curves reached at least 99\% of their respective maximum performance. Quantiles whose maximum performance did not exceed 0.005 were excluded. This criterion captures recovery across the target-performance distribution rather than relying on mean performance alone. Additional motivation and threshold-sensitivity analyses are provided in Appendix A.3.

\subsection{Generalization analyses}

To assess cross-participant generalization, we held out one participant at a time and learned an RRR subspace from the remaining participants (figure \ref{fig:methods}d). We transferred this subspace to the held-out participant and fit a new encoding model using only that participant's neural responses. We then compared transferred RRR with RRR learned from the held-out participant and with PCA across ranks (figure \ref{fig:methods}e).

For cross-dataset transfer, we learned an RRR or PCA subspace from a source dataset and projected the target dataset's embeddings into that space. A new ridge model was then fit and evaluated using only target-dataset neural responses. Thus, in both generalization analyses, only the embedding subspace was transferred; the mapping from that subspace to neural activity was estimated separately in the target participant or dataset.

\subsection{Language-feature decoding}

To characterize the information preferentially represented in brain-predictive subspaces, we annotated stimulus words with lexical, syntactic, semantic, discourse, timing, and pause features. At each rank, we projected embeddings into the RRR and PCA spaces and trained held-out decoders for each available feature. Continuous features were evaluated using Pearson correlation and categorical features using balanced accuracy. The complete feature set and decoding details are provided in Appendix A.7.

\section{Results}

\subsection{Alignment Relies on a Low-Rank Subspace}

\begin{figure}[t]
\begin{center}
\includegraphics[width=\textwidth]{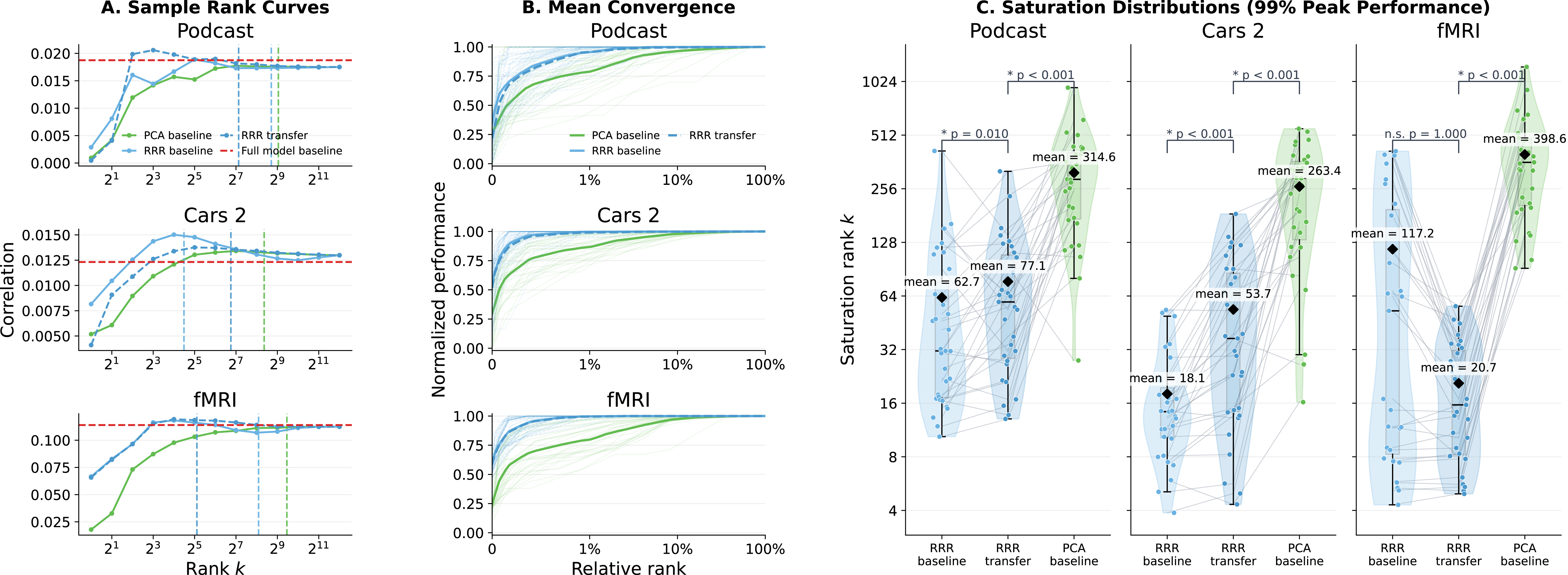}
\end{center}
\caption{\textbf{Brain-predictive subspaces generalize across held-out participants.}
\textbf{(a)} Example rank--performance curves for within-participant RRR, transferred RRR learned from the remaining participants, and PCA. \textbf{(b)} Mean convergence curves across models and layers. \textbf{(c)} Saturation ranks for matched model--layer combinations. Transferred RRR reaches near-maximal performance at substantially lower rank than PCA and is generally comparable to within-participant RRR.}
\label{fig:loso}
\end{figure}
To estimate the dimensionality of the LLM representation required to recover its full neural predictivity, we fit ridge reduced-rank regression (RRR) in each of our three datasets and measured encoding performance as we varied the rank of the predictive bottleneck (Figure \ref{fig:low}; see Appendix for full encoding-model results). Across datasets, models, and layers, performance typically peaked at ranks substantially below the full embedding dimensionality, with low-rank RRR models often outperforming the corresponding full-space ridge baseline (Figure \ref{fig:low}a). This improvement was not observed in PCA-based dimensionality reduction, indicating that it is not simply a consequence of compressing the embedding space. Instead, the result is consistent with RRR preferentially retaining directions that are informative for predicting neural responses while discarding dimensions that contribute little useful predictive signal. Moreover, RRR saturates at only tens of dimensions across all three datasets and across model layers—for example, a mean rank of 20 in the Podcast dataset—whereas PCA generally requires hundreds of dimensions to achieve comparable performance (Figure \ref{fig:low}c). Thus, the directions that support neural prediction constitute a much lower-dimensional subspace than would be inferred from the dominant axes of variance in the embeddings themselves. This low-rank finding is consistent across models, layers, and recording modalities. Whether the same brain-predictive subspaces generalize across participants, stimuli, and datasets, however, remains an open question addressed in the following analyses.

\subsection{The Learned Low-Rank Subspace Generalizes Across Held-Out Subjects}

We next asked whether the brain-predictive subspace learned from one group of participants generalizes to an unseen participant (Figure \ref{fig:methods}d). If the same low-dimensional embedding directions support neural prediction in a held-out participant, this would indicate that the learned subspace reflects structure shared across individuals rather than idiosyncratic properties of a particular participant's neural responses.

Across all three datasets, transferred RRR subspaces recovered near-maximal neural prediction using only tens of dimensions (Figure \ref{fig:loso}). The rank required to reach 99\% of peak performance was substantially lower for transferred RRR subspaces than for PCA across datasets, indicating that the cross-participant result cannot be explained simply by retaining the dominant dimensions of variation in the embeddings. We additionally compared transferred subspaces with RRR subspaces learned using the held-out participant's responses. In the Podcast and fMRI datasets, the transferred subspaces required a comparable number of dimensions—and in some cases fewer—to reach saturation, whereas the Cars 2 dataset showed a larger difference between transferred and participant-inclusive RRR. These results indicate that much of the low-dimensional structure supporting LLM–brain alignment is shared across participants and can be identified without observing neural responses from the participant on whom it is subsequently evaluated.

More broadly, the ability to transfer these subspaces across participants motivates a stronger test of their generality: whether the same predictive directions also transfer across datasets, stimuli, and neural recording modalities.

\subsection{Cross-Dataset Transfer Reveals Shared Brain-Predictive Structure but Is Limited by Stimulus Shift}

\begin{figure}[t]
\begin{center}
\includegraphics[width=\textwidth]{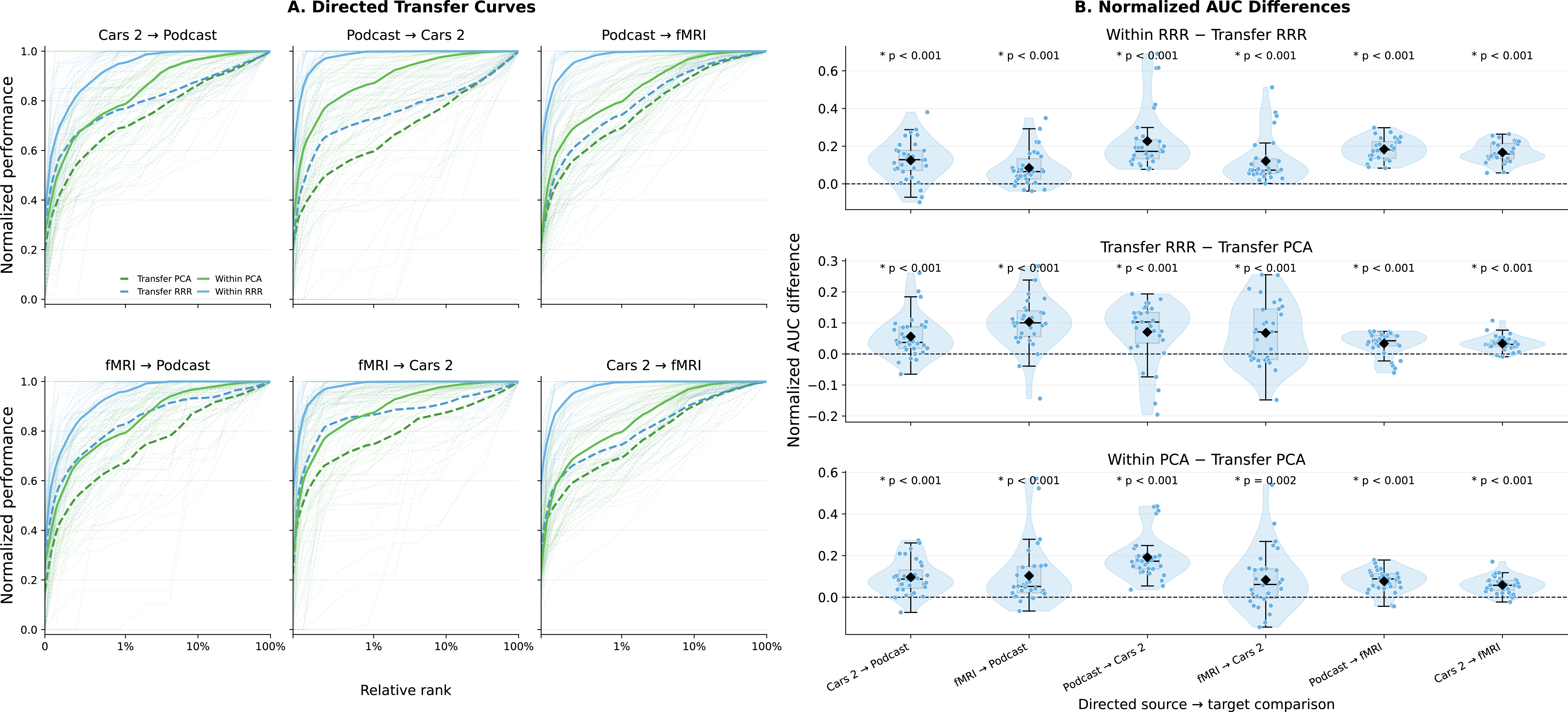}
\end{center}
\caption{\textbf{Brain-predictive subspaces partially generalize across datasets.}
\textbf{(a)} Mean convergence curves for within-dataset RRR, cross-dataset transferred RRR, and transferred PCA. \textbf{(b)} Summary of rank--performance efficiency across directed source--target dataset pairs. Transferred RRR is less efficient than within-dataset RRR but consistently more efficient than transferred PCA, indicating partial cross-dataset generalization.}
\label{cross}
\end{figure}

We next asked whether brain-predictive subspaces generalize across datasets differing in participants, stimuli, recording modality, and preprocessing. For each directed source--target pair, we learned an RRR subspace from the source dataset, transferred it to the target embeddings, and fit a new ridge model using only target neural responses such that only the embedding subspace was transferred. Reduced transfer performance could therefore reflect either differences in neurally predictive directions across datasets or differences in the stimulus distributions used to estimate those directions.

Transferred RRR remained predictive but recovered performance less efficiently than RRR learned within the target dataset (Figure \ref{cross}a). Because these transferred curves saturated more gradually, we summarized normalized rank--performance curves using area under the curve (AUC), with larger AUC indicating more efficient recovery of predictive performance. Across directed dataset pairs and model layers, transferred RRR had lower AUC than within-dataset RRR but significantly higher AUC than transferred PCA for every dataset pair (Figure \ref{cross}b). Thus, brain-predictive subspaces contain structure that generalizes across datasets, although this structure is not fully invariant to changes in stimulus and recording setting.

We next asked whether some of this RRR transfer loss could be explained by differences in the stimuli themselves. To obtain a neural-response-independent measure of cross-dataset representational shift, we performed the same transfer analysis using PCA, which is estimated solely from the stimulus embeddings. The difference between within-dataset and transferred PCA therefore quantifies how much the dominant embedding geometry changes across stimulus sets. Transferred PCA performed significantly worse than within-dataset PCA across all directed dataset comparisons (Figure \ref{cross}b), indicating substantial stimulus-dependent variation in embedding geometry.

\begin{figure}[h]
\begin{center}
\includegraphics[width=0.5\textwidth]{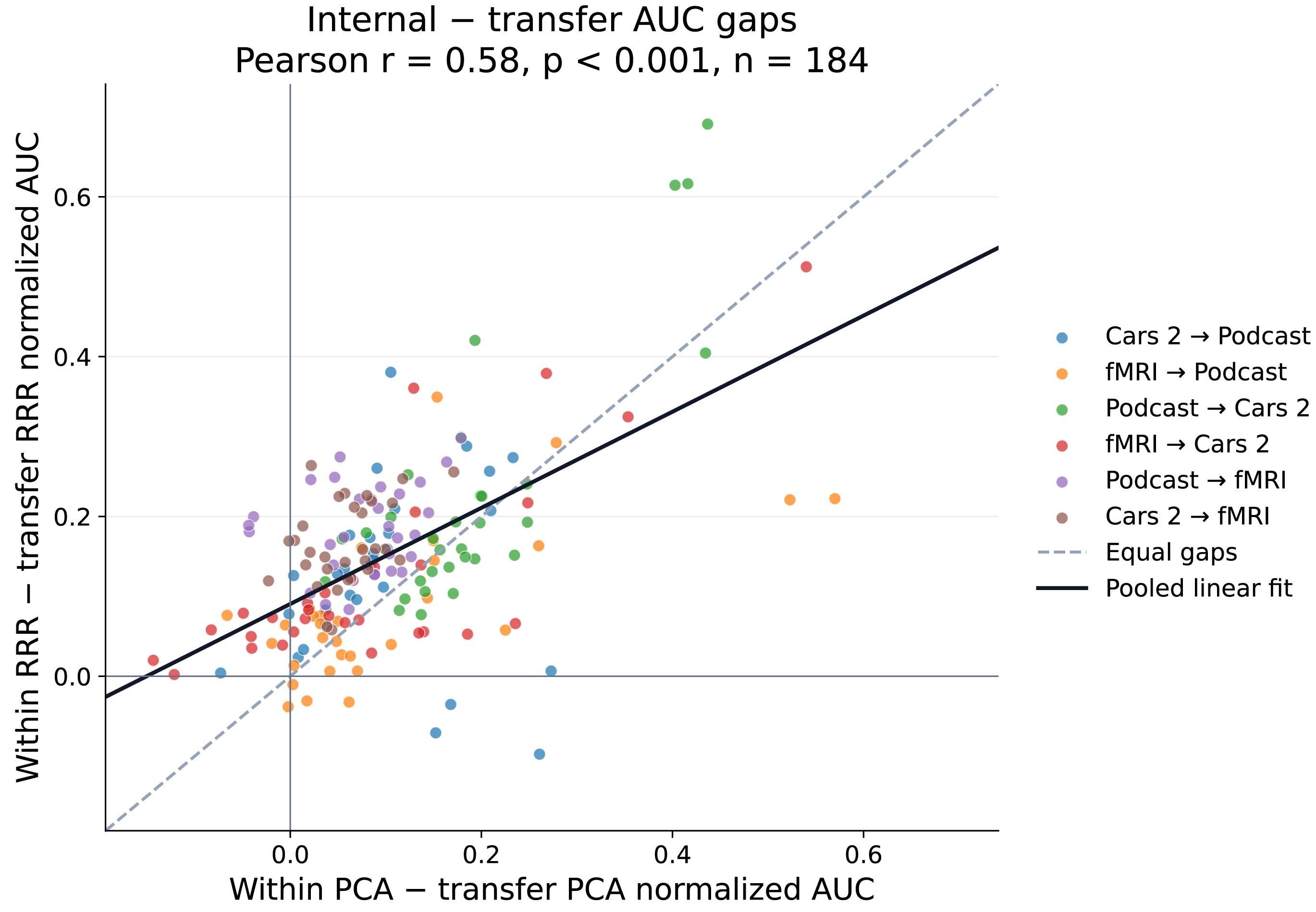}
\end{center}
\caption{\textbf{Cross-dataset RRR transfer loss covaries with stimulus-dependent PCA shift.}
Across models, layers, and directed dataset pairs, the gap between within-dataset and transferred RRR performance is positively correlated with the corresponding PCA transfer gap ($r=0.58$, $n=184$). Each point represents a model--layer combination for one directed source--target transfer.}
\label{corr}
\end{figure}

If stimulus distribution shift contributes to the loss of RRR transfer performance, then dataset pairs with a larger PCA transfer gap should also tend to have a larger RRR transfer gap. We therefore compared these two quantities across models, layers, and directed dataset pairs. The RRR transfer gap positively covaried with the corresponding PCA transfer gap (Figure \ref{corr}; Pearson $r=0.58$, $n=184$), indicating that poorer transfer of the stimulus-defined embedding space was associated with poorer transfer of the brain-predictive subspace.

This relationship does not establish that stimulus shift fully explains the RRR transfer gap, since recording modality, preprocessing, or dataset-specific neural representations may also contribute. However, it indicates that stimulus distribution shift is an important source of cross-dataset variability. We next examine these representational differences more directly using interpretable feature decoding.

\subsection{FMRI and IEEG Aligned Low-Rank Subspaces Emphasize Different Linguistic Information}

Having shown that RRR recovers neural predictivity in far fewer dimensions than PCA, we next asked what information distinguishes these two low-dimensional spaces. PCA preserves directions of greatest variance in the model embeddings, whereas RRR selects directions specifically for their ability to predict neural responses. Comparing feature decodability between equally sized RRR and PCA subspaces therefore asks which linguistic information is preferentially preserved when the embedding space is constrained by neural prediction rather than by variance alone. Greater decodability from RRR does not imply that a feature itself drives neural prediction, but indicates enrichment of that feature, or correlated information, in the brain-predictive subspace.

\begin{figure}[t!]
\begin{center}
\includegraphics[width=\textwidth]{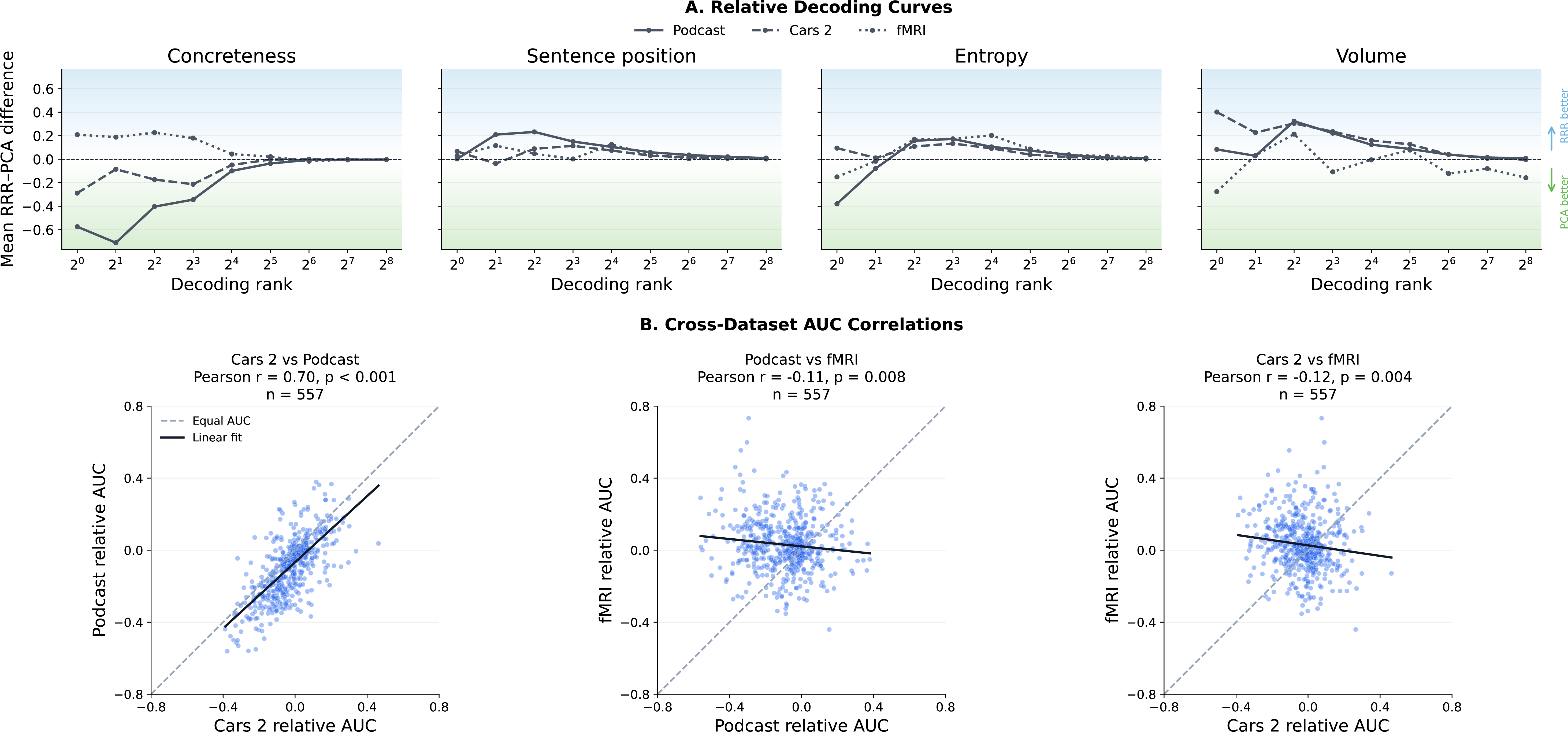}
\end{center}
\caption{\textbf{Linguistic feature enrichment differs across neural datasets.}
(a) Relative decoding advantage of RRR over PCA across ranks for example features. Positive values indicate greater decodability from RRR; negative values indicate greater decodability from PCA. Details of computation in A.3.1. (b) Feature-enrichment profiles are strongly correlated between the two iEEG datasets but not between either iEEG dataset and fMRI.}
\label{decode}
\end{figure}

Across datasets, several features showed preferential decoding from the RRR subspace. For example, sentence position and GPT-2 next-word prediction entropy were more accessible from RRR than PCA at low ranks across all three datasets (Figure \ref{decode}a). Word concreteness, however, was preferentially represented in the fMRI-derived RRR subspace at low ranks, but showed no corresponding enrichment in either iEEG dataset. Thus, the learned subspaces preserve both shared and dataset-specific linguistic information.

We next asked whether these dataset differences showed systematic structure. If RRR is capturing meaningful properties of the neural measurements rather than an arbitrary predictive compression, datasets sharing a recording modality should prioritize similar features. Consistent with this prediction, enrichment profiles were strongly correlated between the two iEEG datasets (Podcast versus Cars 2: ($r=0.70$); Figure \ref{decode}b), but showed little correspondence between either iEEG dataset and fMRI (Podcast versus fMRI: ($r=-0.11$); Cars 2 versus fMRI: ($r=-0.12$)).

This pattern provides convergent support for the interpretation that RRR is isolating meaningful brain-predictive structure. Despite differences in participants and stimuli, the two iEEG datasets emphasize similar linguistic information, while both differ from fMRI. It also complements the cross-dataset transfer results: stimulus shift contributes to transfer loss, but datasets may additionally differ in which components of the embedding space are selected by neural supervision. 

\section{Discussion}

Our results show that LLM--brain alignment during naturalistic language processing is supported by a surprisingly low-dimensional subset of the embedding space. Across datasets, models, and layers, neural prediction saturated using only tens of dimensions, and these brain-predictive directions differed from the principal axes of embedding variance. The learned subspaces generalized strongly across held-out participants and partially across datasets, with cross-dataset differences associated with both stimulus-dependent shifts in embedding geometry and differences in the linguistic information emphasized across recording modalities.

Beyond establishing the existence and generality of this low-rank structure, the framework provides a way to probe which components of large model representations contribute most strongly to neural prediction. Because RRR identifies directions that are selected specifically for their ability to predict neural responses, comparing the information available in the RRR and PCA subspaces can reveal representational mismatches between the two spaces. For example, we find that some stimulus-level properties, including volume (Figure \ref{decode}a), can be decoded more accurately from the RRR space than from a comparable PCA space. This should not be interpreted as evidence that the language model represents acoustic volume: the model receives only the transcript. Instead, the RRR subspace is likely preserving linguistic features that covary with acoustic properties of the stimulus which is useful for predicting responses to heard speech. This example highlights a broader interpretational challenge for neural encoding models built from high-dimensional language representations. Linguistic, semantic, syntactic, discourse, and acoustic properties are highly correlated in naturalistic stimuli, such that successful neural prediction does not by itself identify which underlying feature is causally responsible for alignment.

An open question remains as to what is limiting the dimensionality of this aligned subspace. One possibility is that coarse neural recordings which we typically have access to in human neuroscience contain little signal and we are in fact recovering it well with tens of dimensions. Another possibility though is that the model embedding representations are missing some portion of the true representational signal contained in this neural data. While our analysis cannot answer this question, it does provide evidence that the primary features of LLM embeddings which align with the neural data are not equivalent to the principal axes of the embeddings, demonstrating a divide in the representational geometry of the two systems.

\subsection*{AI use statement}

In this work, we used generative AI tools to help provide feedback on research methodology, implement methods through the writing of code, clean and reformat datasets through scripts for one-off runs, and, at times, interpret results.

We did not use generative AI tools to help develop theoretical models or conceptual frameworks, formulate mathematical claims, propose or refine hypotheses, or support qualitative and thematic data analysis. Generation of synthetic datasets, assistance with translation, assistance in writing proofs, and providing critical ingredients for or proving mathematical claims were not applicable to this work.

Additionally, we used generative AI tools to help create scientific figures or images, suggest experimental parameters, create or edit software code with human review, summarize existing literature, brainstorm, edit our paper to improve readability, and identify relevant literature. Most of the AI usage for this project was to aid in writing code, always with human review. It was also used frequently to make language more concise to fit within the page limitations. We have reviewed all AI-assisted work. We take responsibility for the final content of this work, including text, claims, or artifacts produced with the aid of generative AI.

\subsection*{Reproducibility statement}

To make our work easily reproducible we have provided our code at \href{https://github.com/thefinnlab/low-rank-alignment}{https://github.com/thefinnlab/low-rank-alignment}. In our docs folder is a guide for reproducing which specifies all of the steps necessary to rebuild our results from scratch. All of the data is already open-source and free to download and thus can be easily utilized by any researcher. We have also provided descriptions of our methods in detailed prose in the appendix which should be useful for discerning the specifics of our approach at a high level.

\subsubsection*{Acknowledgments}
We thank Clara Sava- Segal, Katie O’Nell, Tommy Botch, and Kay Liang for their thoughtful comments during the drafting process of this paper.

\bibliography{iclr2027_conference}

@inproceedings{antonelloEvidenceFMRISupports2024,
  title = {Evidence from {{fMRI Supports}} a {{Two-Phase Abstraction Process}} in {{Language Models}}},
  booktitle = {{{NeurIPS}} 2024 {{Workshops}}: {{UniReps}}},
  author = {Antonello, Richard and Cheng, Emily},
  year = 2024,
  urldate = {2026-09-25},
  langid = {english}
}

@inproceedings{antonelloLowdimensionalStructureSpace2021,
  title = {Low-Dimensional {{Structure}} in the {{Space}} of {{Language Representations}} Is {{Reflected}} in {{Brain Responses}}},
  booktitle = {Advances in {{Neural Information Processing Systems}}},
  author = {Antonello, Richard and Turek, Javier S and Vo, Vy and Huth, Alexander},
  year = 2021,
  volume = {34},
  pages = {8332--8344},
  publisher = {Curran Associates, Inc.},
  urldate = {2026-09-25}
}

@article{antonelloPredictiveCodingJust2024,
  title = {Predictive {{Coding}} or {{Just Feature Discovery}}? {{An Alternative Account}} of {{Why Language Models Fit Brain Data}}},
  shorttitle = {Predictive {{Coding}} or {{Just Feature Discovery}}?},
  author = {Antonello, Richard and Huth, Alexander},
  year = 2024,
  month = apr,
  journal = {Neurobiology of Language},
  volume = {5},
  number = {1},
  pages = {64--79},
  issn = {2641-4368},
  doi = {10.1162/nol_a_00087},
  urldate = {2026-09-25}
}

@inproceedings{antonelloScalingLawsLanguage2023,
  title = {Scaling Laws for Language Encoding Models in {{fMRI}}},
  booktitle = {Advances in {{Neural Information Processing Systems}}},
  author = {Antonello, Richard and Vaidya, Aditya and Huth, Alexander},
  year = 2023,
  volume = {36},
  pages = {21895--21907},
  publisher = {Curran Associates, Inc.},
  doi = {10.52202/075280-0958},
  urldate = {2026-09-25}
}

@article{brysbaertConcretenessRatings402014,
  title = {Concreteness Ratings for 40 Thousand Generally Known {{English}} Word Lemmas},
  author = {Brysbaert, Marc and Warriner, Amy Beth and Kuperman, Victor},
  year = 2014,
  month = sep,
  journal = {Behavior Research Methods},
  volume = {46},
  number = {3},
  pages = {904--911},
  issn = {1554-3528},
  doi = {10.3758/s13428-013-0403-5},
  urldate = {2026-09-25},
  langid = {english}
}

@article{brysbaertWordPrevalenceNorms2019,
  title = {Word Prevalence Norms for 62,000 {{English}} Lemmas},
  author = {Brysbaert, Marc and Mandera, Pawe{\l} and McCormick, Samantha F. and Keuleers, Emmanuel},
  year = 2019,
  month = apr,
  journal = {Behavior Research Methods},
  volume = {51},
  number = {2},
  pages = {467--479},
  issn = {1554-3528},
  doi = {10.3758/s13428-018-1077-9},
  urldate = {2026-09-25},
  langid = {english}
}

@article{caucheteuxBrainsAlgorithmsPartially2022,
  title = {Brains and Algorithms Partially Converge in Natural Language Processing},
  author = {Caucheteux, Charlotte and King, Jean-R{\'e}mi},
  year = 2022,
  month = feb,
  journal = {Communications Biology},
  volume = {5},
  number = {1},
  pages = {134},
  publisher = {Nature Publishing Group},
  issn = {2399-3642},
  doi = {10.1038/s42003-022-03036-1},
  urldate = {2026-09-25},
  copyright = {2022 The Author(s)},
  langid = {english}
}

@article{goldsteinSharedComputationalPrinciples2022,
  title = {Shared Computational Principles for Language Processing in Humans and Deep Language Models},
  author = {Goldstein, Ariel and Zada, Zaid and Buchnik, Eliav and Schain, Mariano and Price, Amy and Aubrey, Bobbi and Nastase, Samuel A. and Feder, Amir and Emanuel, Dotan and Cohen, Alon and Jansen, Aren and Gazula, Harshvardhan and Choe, Gina and Rao, Aditi and Kim, Catherine and Casto, Colton and Fanda, Lora and Doyle, Werner and Friedman, Daniel and Dugan, Patricia and Melloni, Lucia and Reichart, Roi and Devore, Sasha and Flinker, Adeen and Hasenfratz, Liat and Levy, Omer and Hassidim, Avinatan and Brenner, Michael and Matias, Yossi and Norman, Kenneth A. and Devinsky, Orrin and Hasson, Uri},
  year = 2022,
  month = mar,
  journal = {Nature Neuroscience},
  volume = {25},
  number = {3},
  pages = {369--380},
  publisher = {Nature Publishing Group},
  issn = {1546-1726},
  doi = {10.1038/s41593-022-01026-4},
  urldate = {2026-09-25},
  copyright = {2022 The Author(s)},
  langid = {english}
}

@article{hadidiSpuriousAlignmentLarge2026,
  title = {Spurious Alignment between Large Language Models and Brains Can Emerge from Non-Robust Methods and Overlooked Confounds},
  author = {Hadidi, Nima and Feghhi, Ebrahim and Song, Bryan H. and Blank, Idan A. and Kao, Jonathan C.},
  year = 2026,
  month = apr,
  journal = {Nature Communications},
  volume = {17},
  number = {1},
  pages = {5769},
  publisher = {Nature Publishing Group},
  issn = {2041-1723},
  doi = {10.1038/s41467-026-72253-7},
  urldate = {2026-09-25},
  copyright = {2026 The Author(s)},
  langid = {english}
}

@article{hoffmanSemanticDiversityMeasure2013,
  title = {Semantic Diversity: {{A}} Measure of Semantic Ambiguity Based on Variability in the Contextual Usage of Words},
  shorttitle = {Semantic Diversity},
  author = {Hoffman, Paul and Lambon Ralph, Matthew A. and Rogers, Timothy T.},
  year = 2013,
  month = sep,
  journal = {Behavior Research Methods},
  volume = {45},
  number = {3},
  pages = {718--730},
  issn = {1554-3528},
  doi = {10.3758/s13428-012-0278-x},
  urldate = {2026-09-25},
  langid = {english}
}

@article{huthNaturalSpeechReveals2016,
  title = {Natural Speech Reveals the Semantic Maps That Tile Human Cerebral Cortex},
  author = {Huth, Alexander G. and {de Heer}, Wendy A. and Griffiths, Thomas L. and Theunissen, Fr{\'e}d{\'e}ric E. and Gallant, Jack L.},
  year = 2016,
  month = apr,
  journal = {Nature},
  volume = {532},
  number = {7600},
  pages = {453--458},
  publisher = {Nature Publishing Group},
  issn = {1476-4687},
  doi = {10.1038/nature17637},
  urldate = {2026-09-25},
  copyright = {2016 Springer Nature Limited},
  langid = {english}
}

@article{izenmanReducedrankRegressionMultivariate1975,
  title = {Reduced-Rank Regression for the Multivariate Linear Model},
  author = {Izenman, Alan Julian},
  year = 1975,
  month = jun,
  journal = {Journal of Multivariate Analysis},
  volume = {5},
  number = {2},
  pages = {248--264},
  issn = {0047-259X},
  doi = {10.1016/0047-259X(75)90042-1},
  urldate = {2026-09-25}
}

@inproceedings{pennington2014glove,
  title     = {{GloVe}: Global Vectors for Word Representation},
  author    = {Pennington, Jeffrey and Socher, Richard and Manning, Christopher D.},
  booktitle = {Proceedings of the 2014 Conference on Empirical Methods in Natural Language Processing (EMNLP)},
  pages     = {1532--1543},
  year      = {2014}
}

@article{hosseiniArtificialNeuralNetwork2024,
  title = {Artificial {{Neural Network Language Models Predict Human Brain Responses}} to {{Language Even After}} a {{Developmentally Realistic Amount}} of {{Training}}},
  author = {Hosseini, Eghbal A. and Schrimpf, Martin and Zhang, Yian and Bowman, Samuel and Zaslavsky, Noga and Fedorenko, Evelina},
  year = 2024,
  month = apr,
  journal = {Neurobiology of Language},
  volume = {5},
  number = {1},
  pages = {43--63},
  issn = {2641-4368},
  doi = {10.1162/nol_a_00137},
  urldate = {2026-09-26},
  pmcid = {PMC11025646},
  pmid = {38645622}
}

@misc{liuRoBERTaRobustlyOptimized2019,
  title = {{{RoBERTa}}: {{A Robustly Optimized BERT Pretraining Approach}}},
  shorttitle = {{{RoBERTa}}},
  author = {Liu, Yinhan and Ott, Myle and Goyal, Naman and Du, Jingfei and Joshi, Mandar and Chen, Danqi and Levy, Omer and Lewis, Mike and Zettlemoyer, Luke and Stoyanov, Veselin},
  year = 2019,
  month = jul,
  number = {arXiv:1907.11692},
  eprint = {1907.11692},
  primaryclass = {cs.CL},
  publisher = {arXiv},
  doi = {10.48550/arXiv.1907.11692},
  urldate = {2026-09-26},
  archiveprefix = {arXiv}
}

@misc{llama3modelcard,
    title={Llama 3 Model Card},
    author={AI@Meta},
    year={2024},
    url = {https://github.com/meta-llama/llama3/blob/main/MODEL_CARD.md}
}

@misc{gpt-j,
  author = {Wang, Ben and Komatsuzaki, Aran},
  title = {{GPT-J-6B: A 6 Billion Parameter Autoregressive Language Model}},
  howpublished = {\url{https://github.com/kingoflolz/mesh-transformer-jax}},
  year = 2021,
  month = May
}

@inproceedings{jainIncorporatingContextLanguage2018,
  title = {Incorporating {{Context}} into {{Language Encoding Models}} for {{fMRI}}},
  booktitle = {Advances in {{Neural Information Processing Systems}}},
  author = {Jain, Shailee and Huth, Alexander},
  year = 2018,
  volume = {31},
  publisher = {Curran Associates, Inc.},
  urldate = {2026-09-25}
}

@article{kumarSharedFunctionalSpecialization2024,
  title = {Shared Functional Specialization in Transformer-Based Language Models and the Human Brain},
  author = {Kumar, Sreejan and Sumers, Theodore R. and Yamakoshi, Takateru and Goldstein, Ariel and Hasson, Uri and Norman, Kenneth A. and Griffiths, Thomas L. and Hawkins, Robert D. and Nastase, Samuel A.},
  year = 2024,
  month = jun,
  journal = {Nature Communications},
  volume = {15},
  number = {1},
  pages = {5523},
  publisher = {Nature Publishing Group},
  issn = {2041-1723},
  doi = {10.1038/s41467-024-49173-5},
  urldate = {2026-09-25},
  copyright = {2024 The Author(s)},
  langid = {english}
}

@article{kupermanAgeofacquisitionRatings300002012,
  title = {Age-of-Acquisition Ratings for 30,000 {{English}} Words},
  author = {Kuperman, Victor and {Stadthagen-Gonzalez}, Hans and Brysbaert, Marc},
  year = 2012,
  month = dec,
  journal = {Behavior Research Methods},
  volume = {44},
  number = {4},
  pages = {978--990},
  issn = {1554-3528},
  doi = {10.3758/s13428-012-0210-4},
  urldate = {2026-09-25},
  langid = {english}
}

@article{lebelNaturalLanguageFMRI2023,
  title = {A Natural Language {{fMRI}} Dataset for Voxelwise Encoding Models},
  author = {LeBel, Amanda and Wagner, Lauren and Jain, Shailee and {Adhikari-Desai}, Aneesh and Gupta, Bhavin and Morgenthal, Allyson and Tang, Jerry and Xu, Lixiang and Huth, Alexander G.},
  year = 2023,
  month = aug,
  journal = {Scientific Data},
  volume = {10},
  number = {1},
  pages = {555},
  publisher = {Nature Publishing Group},
  issn = {2052-4463},
  doi = {10.1038/s41597-023-02437-z},
  urldate = {2026-09-25},
  copyright = {2023 The Author(s)},
  langid = {english}
}

@article{lewisLowRankTensorEncoding,
  title = {Low-{{Rank Tensor Encoding Models Decompose Natural Speech Comprehension Processes}}},
  year = 2025,
  doi     = {10.1101/2025.06.02.657514},
  journal = {bioRxiv},
  month   = jun,
  author = {Lewis, Lane and Pitkow, Xaq and Wehbe, Leila},
  langid = {english}
}

@article{lynottLancasterSensorimotorNorms2020,
  title = {The {{Lancaster Sensorimotor Norms}}: Multidimensional Measures of Perceptual and Action Strength for 40,000 {{English}} Words},
  shorttitle = {The {{Lancaster Sensorimotor Norms}}},
  author = {Lynott, Dermot and Connell, Louise and Brysbaert, Marc and Brand, James and Carney, James},
  year = 2020,
  month = jun,
  journal = {Behavior Research Methods},
  volume = {52},
  number = {3},
  pages = {1271--1291},
  issn = {1554-3528},
  doi = {10.3758/s13428-019-01316-z},
  urldate = {2026-09-25},
  langid = {english}
}

@article{mischlerContextualFeatureExtraction2024,
  title = {Contextual Feature Extraction Hierarchies Converge in Large Language Models and the Brain},
  author = {Mischler, Gavin and Li, Yinghao Aaron and Bickel, Stephan and Mehta, Ashesh D. and Mesgarani, Nima},
  year = 2024,
  month = dec,
  journal = {Nature Machine Intelligence},
  volume = {6},
  number = {12},
  pages = {1467--1477},
  publisher = {Nature Publishing Group},
  issn = {2522-5839},
  doi = {10.1038/s42256-024-00925-4},
  urldate = {2026-09-25},
  copyright = {2024 The Author(s), under exclusive licence to Springer Nature Limited},
  langid = {english}
}

@misc{mohammadNRCVADLexicon2025,
  title = {{{NRC VAD Lexicon}} v2: {{Norms}} for {{Valence}}, {{Arousal}}, and {{Dominance}} for over 55k {{English Terms}}},
  shorttitle = {{{NRC VAD Lexicon}} V2},
  author = {Mohammad, Saif M.},
  year = 2025,
  month = mar,
  number = {arXiv:2503.23547},
  eprint = {2503.23547},
  primaryclass = {cs.CL},
  publisher = {arXiv},
  doi = {10.48550/arXiv.2503.23547},
  urldate = {2026-09-25},
  archiveprefix = {arXiv}
}

@article{schrimpfNeuralArchitectureLanguage2021,
  title = {The Neural Architecture of Language: {{Integrative}} Modeling Converges on Predictive Processing},
  shorttitle = {The Neural Architecture of Language},
  author = {Schrimpf, Martin and Blank, Idan Asher and Tuckute, Greta and Kauf, Carina and Hosseini, Eghbal A. and Kanwisher, Nancy and Tenenbaum, Joshua B. and Fedorenko, Evelina},
  year = 2021,
  month = nov,
  journal = {Proceedings of the National Academy of Sciences},
  volume = {118},
  number = {45},
  pages = {e2105646118},
  publisher = {Proceedings of the National Academy of Sciences},
  doi = {10.1073/pnas.2105646118},
  urldate = {2026-09-25}
}

@inproceedings{tonevaInterpretingImprovingNaturallanguage2019,
  title = {Interpreting and Improving Natural-Language Processing (in Machines) with Natural Language-Processing (in the Brain)},
  booktitle = {Advances in {{Neural Information Processing Systems}}},
  author = {Toneva, Mariya and Wehbe, Leila},
  year = 2019,
  volume = {32},
  publisher = {Curran Associates, Inc.},
  urldate = {2026-09-25}
}

@article{wangBrainTreebankLargescale2024,
  title = {Brain {{Treebank}}: {{Large-scale}} Intracranial Recordings from Naturalistic Language Stimuli},
  shorttitle = {Brain {{Treebank}}},
  author = {Wang, Christopher and Yaari, Adam and Singh, Aaditya K. and Subramaniam, Vighnesh and Rosenfarb, Dana and DeWitt, Jan and Misra, Pranav and Madsen, Joseph R. and Stone, Scellig and Kreiman, Gabriel and Katz, Boris and Cases, Ignacio and Barbu, Andrei},
  year = 2024,
  month = dec,
  journal = {Advances in Neural Information Processing Systems},
  volume = {37},
  pages = {96505--96540},
  doi = {10.52202/079017-3060},
  urldate = {2026-09-25},
  langid = {english}
}

@article{zadaPodcastECoGDataset2025,
  title = {The ``{{Podcast}}'' {{ECoG}} Dataset for Modeling Neural Activity during Natural Language Comprehension},
  author = {Zada, Zaid and Nastase, Samuel A. and Aubrey, Bobbi and Jalon, Itamar and Michelmann, Sebastian and Wang, Haocheng and Hasenfratz, Liat and Doyle, Werner and Friedman, Daniel and Dugan, Patricia and Melloni, Lucia and Devore, Sasha and Flinker, Adeen and Devinsky, Orrin and Goldstein, Ariel and Hasson, Uri},
  year = 2025,
  month = jul,
  journal = {Scientific Data},
  volume = {12},
  number = {1},
  pages = {1135},
  publisher = {Nature Publishing Group},
  issn = {2052-4463},
  doi = {10.1038/s41597-025-05462-2},
  urldate = {2026-09-25},
  copyright = {2025 The Author(s)},
  langid = {english}
}

@article{dedeyneSmallWorldWords2019,
  title = {The ``{{Small World}} of {{Words}}'' {{English}} Word Association Norms for over 12,000 Cue Words},
  author = {De Deyne, Simon and Navarro, Danielle J. and Perfors, Amy and Brysbaert, Marc and Storms, Gert},
  year = 2019,
  month = jun,
  journal = {Behavior Research Methods},
  volume = {51},
  number = {3},
  pages = {987--1006},
  issn = {1554-3528},
  doi = {10.3758/s13428-018-1115-7},
  urldate = {2026-09-26},
  langid = {english}
}

@article{shannonMathematicalTheoryCommunication1948,
  title = {A Mathematical Theory of Communication},
  author = {Shannon, C. E.},
  year = 1948,
  month = jul,
  journal = {The Bell System Technical Journal},
  volume = {27},
  number = {3},
  pages = {379--423},
  issn = {0005-8580},
  doi = {10.1002/j.1538-7305.1948.tb01338.x},
  urldate = {2026-09-26}
}

@inproceedings{nivreUniversalDependenciesV22020,
  title = {Universal {{Dependencies}} v2: {{An Evergrowing Multilingual Treebank Collection}}},
  shorttitle = {Universal {{Dependencies}} V2},
  booktitle = {Proceedings of the {{Twelfth Language Resources}} and {{Evaluation Conference}}},
  author = {Nivre, Joakim and {de Marneffe}, Marie-Catherine and Ginter, Filip and Haji{\v c}, Jan and Manning, Christopher D. and Pyysalo, Sampo and Schuster, Sebastian and Tyers, Francis and Zeman, Daniel},
  editor = {Calzolari, Nicoletta and B{\'e}chet, Fr{\'e}d{\'e}ric and Blache, Philippe and Choukri, Khalid and Cieri, Christopher and Declerck, Thierry and Goggi, Sara and Isahara, Hitoshi and Maegaard, Bente and Mariani, Joseph and Mazo, H{\'e}l{\`e}ne and Moreno, Asuncion and Odijk, Jan and Piperidis, Stelios},
  year = 2020,
  month = may,
  pages = {4034--4043},
  publisher = {European Language Resources Association},
  address = {Marseille, France},
  urldate = {2026-09-26},
  isbn = {979-10-95546-34-4},
  langid = {english}
}

@inproceedings{qiStanzaPythonNatural2020,
  title = {Stanza: {{A Python Natural Language Processing Toolkit}} for {{Many Human Languages}}},
  shorttitle = {Stanza},
  booktitle = {Proceedings of the 58th {{Annual Meeting}} of the {{Association}} for {{Computational Linguistics}}: {{System Demonstrations}}},
  author = {Qi, Peng and Zhang, Yuhao and Zhang, Yuhui and Bolton, Jason and Manning, Christopher D.},
  editor = {Celikyilmaz, Asli and Wen, Tsung-Hsien},
  year = 2020,
  month = jul,
  pages = {101--108},
  publisher = {Association for Computational Linguistics},
  address = {Online},
  doi = {10.18653/v1/2020.acl-demos.14},
  urldate = {2026-09-26}
}

@article{gibsonLinguisticComplexityLocality1998,
  title = {Linguistic Complexity: Locality of Syntactic Dependencies},
  shorttitle = {Linguistic Complexity},
  author = {Gibson, Edward},
  year = 1998,
  month = aug,
  journal = {Cognition},
  volume = {68},
  number = {1},
  pages = {1--76},
  issn = {0010-0277},
  doi = {10.1016/S0010-0277(98)00034-1},
  urldate = {2026-09-26}
}

@inproceedings{rollPSSTProsodicSpeech2023,
  title = {{{PSST}}! {{Prosodic Speech Segmentation}} with {{Transformers}}},
  booktitle = {Proceedings of the 27th {{Conference}} on {{Computational Natural Language Learning}} ({{CoNLL}})},
  author = {Roll, Nathan and Graham, Calbert and Todd, Simon},
  editor = {Jiang, Jing and Reitter, David and Deng, Shumin},
  year = 2023,
  month = dec,
  pages = {476--487},
  publisher = {Association for Computational Linguistics},
  address = {Singapore},
  doi = {10.18653/v1/2023.conll-1.31},
  urldate = {2026-09-26}
}

@software{speer2022wordfreq,
    author  = {Speer, Robyn},
    title   = {\texttt{wordfreq}: v3.0},
    year    = {2022},
    version = {3.0.2},
    publisher = {Zenodo},
    doi     = {10.5281/zenodo.7199437}
  }

@article{radford2019language,
    author  = {Radford, Alec and Wu, Jeffrey and Child, Rewon and Luan, David
               and Amodei, Dario and Sutskever, Ilya},
    title   = {Language Models Are Unsupervised Multitask Learners},
    journal = {OpenAI Technical Report},
    year    = {2019},
    url     = {https://cdn.openai.com/better-language-models/language_models_are_unsupervised_multitask_learners.pdf}
  }

@misc{bredinPyannoteaudioNeuralBuilding2019,
  title = {Pyannote.Audio: Neural Building Blocks for Speaker Diarization},
  shorttitle = {Pyannote.Audio},
  author = {Bredin, Herv{\'e} and Yin, Ruiqing and Coria, Juan Manuel and Gelly, Gregory and Korshunov, Pavel and Lavechin, Marvin and Fustes, Diego and Titeux, Hadrien and Bouaziz, Wassim and Gill, Marie-Philippe},
  year = 2019,
  month = nov,
  number = {arXiv:1911.01255},
  eprint = {1911.01255},
  primaryclass = {eess.AS},
  publisher = {arXiv},
  doi = {10.48550/arXiv.1911.01255},
  urldate = {2026-09-26},
  archiveprefix = {arXiv}
}

@misc{princeton2010wordnet,
    author       = {{Princeton University}},
    title        = {About {WordNet}},
    year         = {2010},
    howpublished = {\url{https://wordnet.princeton.edu/}},
    note         = {WordNet, Princeton University; accessed September 26, 2026}
  }

@article{zadaSharedModelbasedLinguistic2024,
  title = {A Shared Model-Based Linguistic Space for Transmitting Our Thoughts from Brain to Brain in Natural Conversations},
  author = {Zada, Zaid and Goldstein, Ariel and Michelmann, Sebastian and Simony, Erez and Price, Amy and Hasenfratz, Liat and Barham, Emily and Zadbood, Asieh and Doyle, Werner and Friedman, Daniel and Dugan, Patricia and Melloni, Lucia and Devore, Sasha and Flinker, Adeen and Devinsky, Orrin and Nastase, Samuel A. and Hasson, Uri},
  year = 2024,
  month = sep,
  journal = {Neuron},
  volume = {112},
  number = {18},
  pages = {3211-3222.e5},
  publisher = {Elsevier},
  issn = {0896-6273},
  doi = {10.1016/j.neuron.2024.06.025},
  urldate = {2026-09-25},
  langid = {english},
  pmid = {39096896}
}
\bibliographystyle{iclr2027_conference}

\appendix
\section{Appendix}

\subsection{Data Preprocessing}

\subsubsection{LLM Embeddings}

We generated contextual representations from the stimulus transcripts using GPT-2, GPT-2 XL, GPT-J 6B, Llama 3 8B, and RoBERTa Large. Each
  transcript word was tokenized independently with a leading space, and the resulting subword tokens were concatenated in transcript order. For the causal models,
  we processed overlapping windows of up to 1,023 tokens, advancing by 512 tokens per window. We retained all token representations from the first window and only
  newly introduced token representations from subsequent windows. Native causal attention restricted each representation to the target token and preceding tokens
  within its window. For RoBERTa, we constructed a separate window of up to 511 tokens ending at each target token and retained its final-position representation.
  Thus, although RoBERTa uses bidirectional attention within a window, its extracted representations had no access to subsequent transcript tokens.

  We extracted hidden states at five depths spanning each model: layers 0, 3, 6, 9, and 12 for GPT-2; 0, 12, 24, 36, and 48 for GPT-2 XL; 0, 7, 14, 21, and 28 for
  GPT-J; 0, 8, 16, 24, and 32 for Llama; and 0, 6, 12, 18, and 24 for RoBERTa. Layer 0 denotes the embedding-stage hidden state. Word representations were
  obtained by averaging the hidden states of all subword tokens belonging to that word. Contextual representations were generated before applying the vocabulary
  restrictions used in the neural analyses. Controls comprised a randomly initialized GPT-2 model with initialization seed 42 and static 300-dimensional GloVe
  embeddings.

\subsubsection{Cars 2 Preprocessing}
  
  We analyzed Brain Treebank intracranial recordings from subjects 3, 7, and 10 during viewing of Cars 2. Contacts identified as corrupted
  in the dataset annotations were removed, and the continuous voltage recordings were resampled from 2,048 Hz to 512 Hz. We then despiked each contact by
  identifying samples whose absolute deviation from its median exceeded four times its interquartile range. The artifact mask was expanded using a 100-ms filter
  applied forward and backward, and masked samples were replaced using piecewise cubic Hermite interpolation.

  We applied shaft-Laplacian rereferencing, subtracting the mean voltage of the two immediately adjacent contacts on the same shaft from each contact:

  \begin{align*}
      V_n^{ref}(t) = V_n(t) - \frac{V_{n-1}(t) + V_{n+1}(t)}{2}
  \end{align*}
  
  Only contacts with both consecutively numbered neighbors available after corrupted-contact removal were retained. We applied notch filters at 60, 120, and 180
  Hz with 2-Hz notch widths, followed by a fourth-order Butterworth band-pass filter from 70 to 200 Hz. High-gamma amplitude was defined as the magnitude of the
  analytic signal obtained through the Hilbert transform.

\subsubsection{Word alignment and epoch construction.}
   We reconstructed transcript words by matching the dataset’s transcript entries to the corresponding sentence text,
  recovering punctuation and contractions where applicable. When multiple separately timed entries mapped to a single sentence word, we retained the separately
  timed entries. Each resulting word received a persistent transcript index linking its embedding to its neural epoch.

  Movie timestamps were mapped separately to each subject’s neural recording using the supplied synchronization triggers.
  Words lacking valid onset or offset times, or beginning after the final synchronization trigger, were excluded from neural alignment.

  For each remaining word, we rounded its recording-relative onset to the nearest sample on the 512-Hz preprocessing grid and extracted a high-gamma epoch
  covering [-2,2) seconds relative to onset. When multiple words mapped to the same onset sample, we retained the first; epochs extending beyond the recording
  boundaries were discarded. Each extracted epoch was then independently resampled to 32 Hz using FFT resampling with a Hamming window, yielding 128 samples per
  contact at 31.25-ms intervals. Consequently, every epoch used a sampling grid anchored to its own word onset.

  We stored the retained transcript indices alongside the epochs and used these indices to select the corresponding word embeddings. Analyses combining subjects
  used the intersection of their retained word indices. The iEEG analyses additionally restricted words to those covered by the 300-dimensional GloVe vocabulary,
  ensuring a common vocabulary across model comparisons.

\subsection{Methods Details}

\subsubsection{RRR fitting and rank evaluation}

Within each training partition, we fit a multivariate ridge model from ($X$) to ($Y$), yielding coefficient matrix ($W$) and fitted responses ($\hat{Y}=XW$). After decomposing ($\hat{Y}$), we used its first ($k$) right singular vectors ($V_k$) to construct ($A_k=WV_k$). Before projection, the truncated input factor was orthonormalized using QR decomposition so that differences in column magnitude did not influence the refitted model or comparisons with PCA.

Ranks generally doubled in powers of 2 to the available embedding-dimensionality endpoint. At each rank, a new ridge encoding model was fit after projection. Ridge penalties were selected by inner cross-validation from a 25-value scale-aware grid. Performance was measured as Pearson correlation between predicted and observed held-out responses. For iEEG, correlations across lags were averaged in Fisher-(z) space to obtain per-electrode performance.

\subsubsection{Construction of the fMRI RRR subspace}

For fMRI, the initial RRR fit operated on the complete four-delay finite impulse response design, yielding a separate input-factor block for each delay. We combined these blocks to obtain a (k)-dimensional subspace in the original embedding coordinates. For each of the first (k) RRR components, coefficients were normalized across the four delays and scaled by the corresponding singular value of the ridge model's fitted training responses. This weighting gives components with greater fitted prediction energy greater influence while preserving their distribution across delays.

The four weighted blocks were concatenated, and their leading (k) left singular vectors were used as an orthonormal embedding basis. At each rank, embeddings were projected into this basis, the four-delay design was reconstructed, and a new ridge encoding model was fit for held-out evaluation.

\subsubsection{Language-feature decoding details}

Stimulus words were annotated with the available lexical, syntactic, semantic, discourse, timing, and pause features. At each rank, embeddings were projected separately into the RRR and PCA spaces. Continuous features were decoded using ridge regression and evaluated by Pearson correlation; categorical features were decoded using class-balanced logistic regression and evaluated by balanced accuracy. Features unavailable for a given dataset were omitted from that dataset's analysis. A complete list of decoded features and their sources is provided below.

\subsection{Saturation rank definition}

Our goal was to estimate the minimum rank required to recover near-maximal neural predictive performance. In principle, this could be defined separately for every neural target (voxel or electrode). Individual target-level rank--performance curves are noisy, however, making target-specific saturation estimates unstable. Conversely, defining saturation from only the mean performance across targets can obscure heterogeneity across the neural-target distribution.

We therefore characterized saturation at the level of the target-performance distribution. At each evaluated rank, we computed the 10th, 25th, 50th, 75th, and 90th percentiles of held-out target performance and tracked each quantile across the rank sweep. For each quantile, we defined its maximum as the highest performance attained across evaluated ranks. The overall saturation rank was then the minimum rank at which every informative quantile reached at least 99\% of its respective maximum performance.

Quantile curves whose maximum performance did not exceed 0.005 were excluded because they contained almost no positive predictive performance for which a recovery rank could be meaningfully defined. Requiring all remaining quantiles to reach threshold prevents saturation from being determined solely by high-performing targets or by changes in the mean and instead requires near-maximal recovery across a broad portion of the neural-target distribution.

Because the estimated saturation rank depends on the chosen recovery threshold, we additionally evaluated the sensitivity of our conclusions to alternative thresholds.

\begin{figure}
    \centering
    \includegraphics[width=\linewidth]{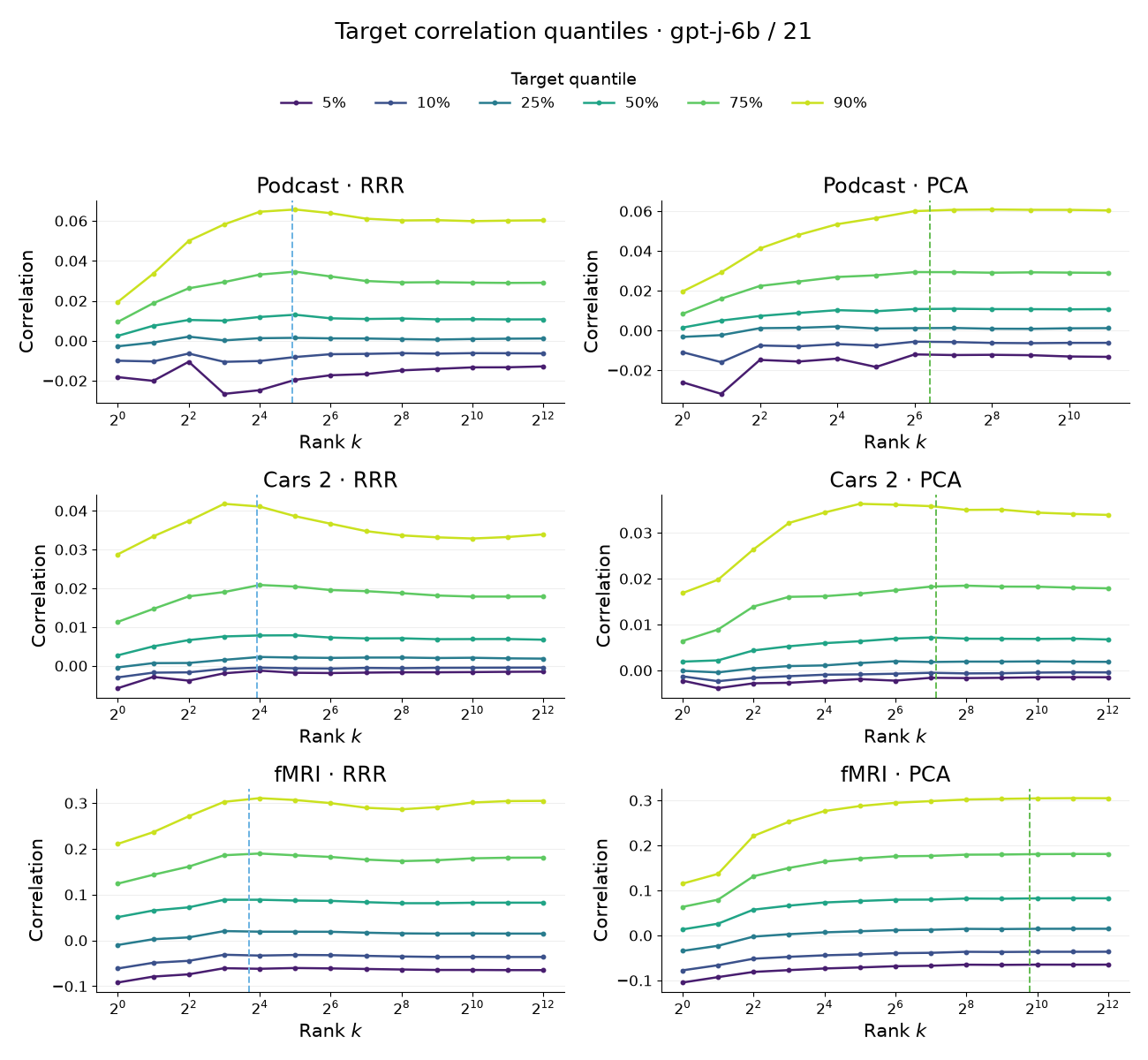}
    \caption{GPT-J-6b layer 21 performance quantiles on different datasets. We can see that lower-quantiles generally lag a bit behind higher ones. We exclude any with a peak value less than 0.005 when calculating saturation to avoid unreliable targets leading to noisy overestimates.}
    \label{fig:quantiles}
\end{figure}

\begin{figure}
    \centering
    \includegraphics[width=\linewidth]{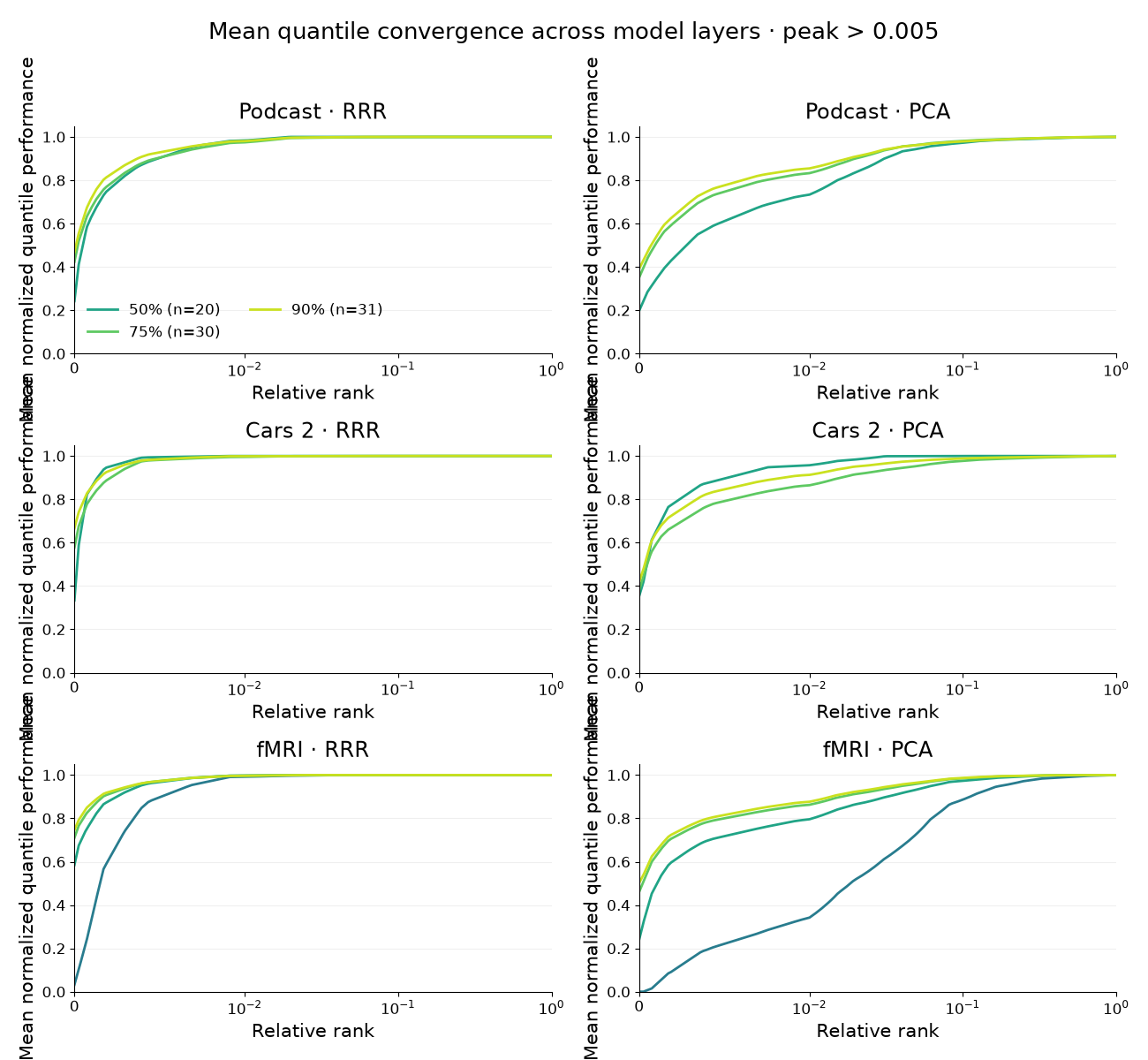}
    \caption{Normalized performance curves for every quantile averaged across model-layers. All tested quantiles tend to converge quickly on RRR but less quickly for PCA.}
    \label{fig:avg-quantiles}
\end{figure}

\begin{figure}
    \centering
    \includegraphics[width=\linewidth]{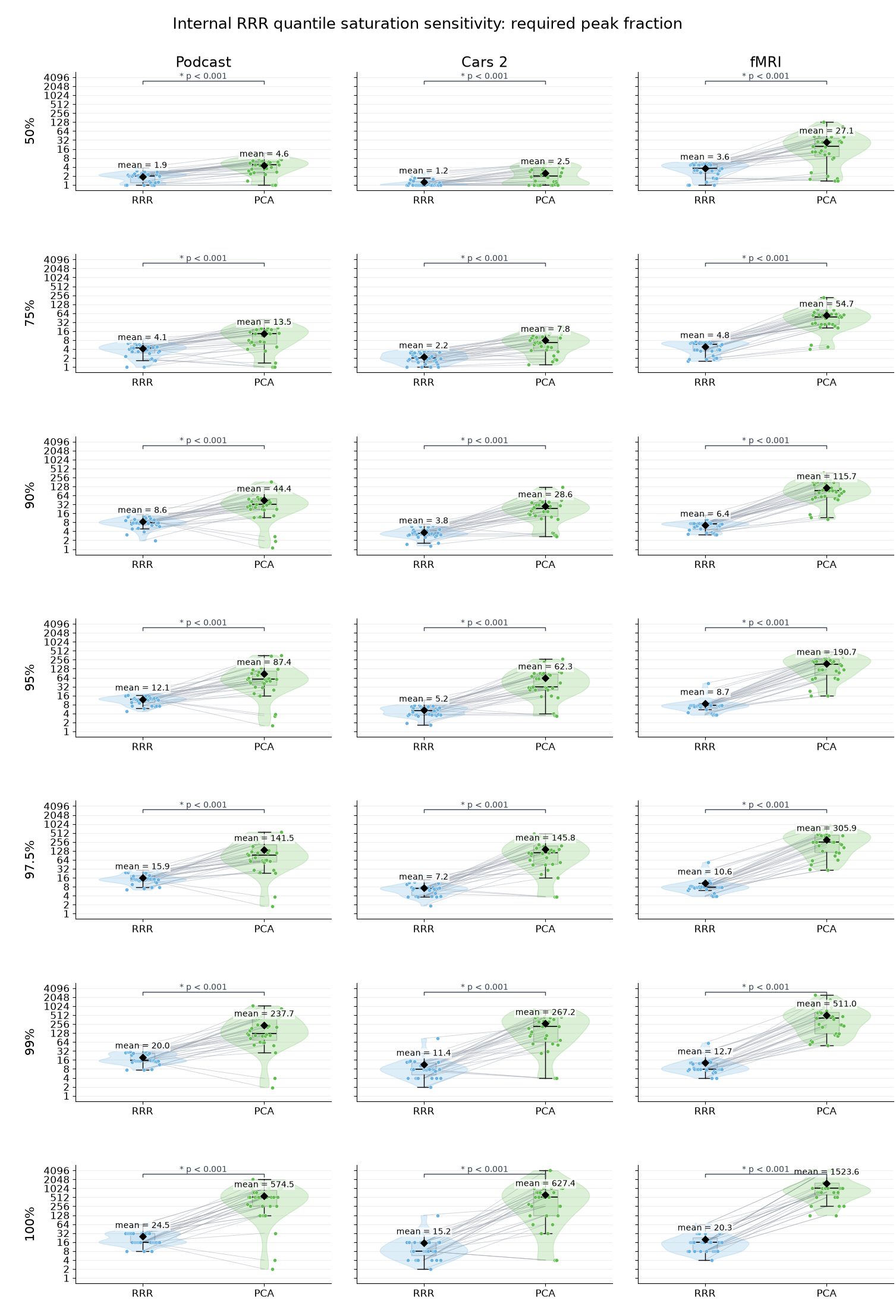}
    \caption{Sensitivity analysis of how saturation rank changes as you change the required performance threshold across qualifying quantiles. Across all thresholds RRR converges quicker than PCA, but the effect becomes more drastic as the value increases. We selected 99\% as our goal is to ask when max performance is almost entirely recovered.}
    \label{fig:perf-sens}
\end{figure}

\begin{figure}
    \centering
    \includegraphics[width=\linewidth]{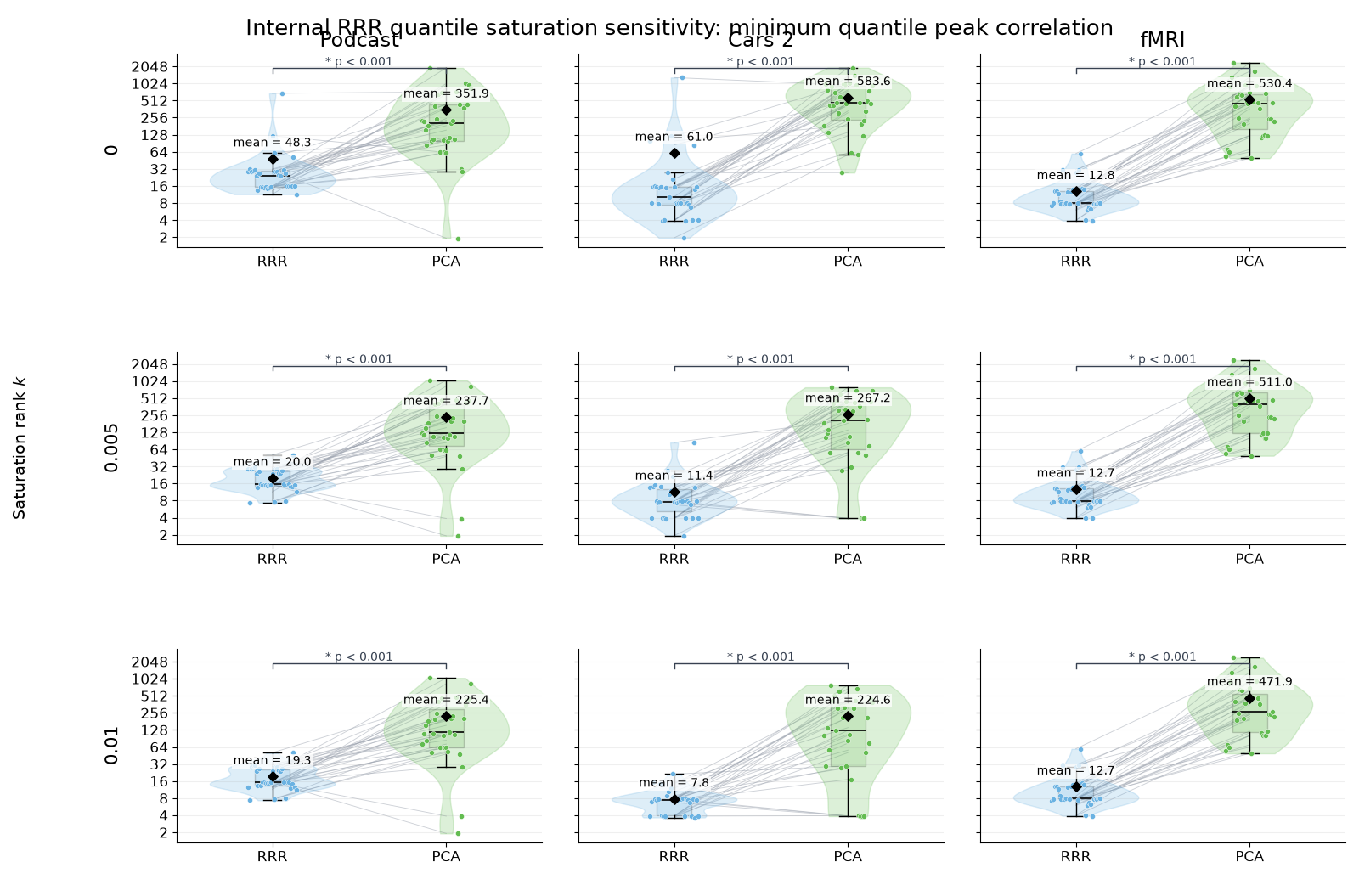}
    \caption{Sensitivity analysis of how saturation rank changes as you change the floor correlation for a quantile to be included in the saturation rank computation. Using a floor of 0 causes some model layers to gain $\sim 1000$ saturation rank. Increasing to 0.005 leads to more reasonable results which we choose as this is so close to a noise response.}
    \label{fig:floor-sens}
\end{figure}

See figures \ref{fig:quantiles}, \ref{fig:avg-quantiles}, \ref{fig:floor-sens}, \ref{fig:perf-sens} for example quantiles and sensitivity analyses of these measures.

\subsubsection{Relative Decoding Performance}

After clipping each decoding score below at zero, we defined relative advantage as
  \begin{align*}
  D(k)=\frac{s_{RRR}(k)-s_{PCA}(k)}{[s_{RRR}(k)+s_{PCA}(k)]/2+0.05}
  \end{align*}

  The value is greater than 0 if a feature can be more readily decoded from the RRR space than the PCA space and less than 0 otherwise. It is normalized by the mean of the two values and 0.05 is added to prevent unstable values at low performance. We repeat this computation across many ranks to study how it changes with the size of the subspaces.

\subsection{Significance Tests}

\subsubsection{Statistical comparisons}

 We compared dimensionality and transfer performance using one-sided paired sign-flip permutation tests. The pairing unit was a model–
  layer combination. For each comparison, the test statistic was the mean paired
  difference. We generated a null distribution by independently multiplying each paired difference by either +1 or -1, with equal probability, across 10,000
  randomizations using a fixed random seed of 0. We calculated Monte Carlo p-values as
  \[
  p = \frac{1 + \Sigma_{b=1}^{10,000} [\text{$T_{b}$ is at least as extreme as $T_{obs}$}]}{10,001}
\]
  where extremeness was defined according to the direction of the one-sided alternative.

\subsubsection{Saturation-rank comparisons.}
   For each dataset, model–layer combination, and method, we summarized dimensionality by the rank required to reach 99\% of peak
  performance. We first averaged target-level correlations across evaluation folds in Fisher-z space and computed the 10th, 25th, 50th, 75th, and 90th
  percentiles across neural targets. Each percentile curve was made monotonic by taking its cumulative maximum and normalized by its own peak. Curves with peak
  correlations at or below 0.005 were excluded. Saturation rank was the largest of the first ranks at which the eligible percentile curves reached the threshold,
  with linear interpolation between tested ranks.

  We tested whether RRR required fewer dimensions than PCA using paired differences in \(\log_2\) saturation rank. For leave-one-subject-out analyses, we computed
  saturation ranks separately for each subject and averaged the finite ranks across subjects within each model–layer combination before applying the logarithmic
  transformation. We tested whether transferred RRR required fewer dimensions than the within-subject PCA baseline and whether the RRR baseline required fewer
  dimensions than transferred RRR. Comparisons used model–layer combinations with finite saturation estimates for every method shown in the panel. The two leave-
  one-subject-out tests were corrected using the Holm procedure within each dataset panel. Sensitivity analyses repeated these comparisons across alternative
  saturation thresholds and, for internal analyses, alternative minimum peak correlations.

\subsubsection{Cross-dataset transfer comparisons}
  We summarized each performance curve by its area under the curve (AUC). Correlations were clipped below at zero, made
  monotonic using a cumulative maximum, and divided by the curve’s own peak. We then integrated the normalized curve using the trapezoidal rule over normalized
  log rank, $log(k)/log(k_{max})$, from rank 1 to the maximum tested rank. For each of the six directed dataset pairs, we tested whether the mean paired AUC difference
  was positive for three comparisons: within-dataset RRR versus transferred RRR, transferred RRR versus transferred PCA, and within-dataset PCA versus transferred
  PCA. Each test used matched model–layer combinations with valid AUCs.

\subsubsection{Correlation analyses}
  We used Pearson correlations with two-sided p-values to assess associations between the within-dataset-minus-transfer AUC gaps for PCA and
  RRR, pooling matched observations across all six transfer directions. We also assessed cross-dataset consistency of the relative decoding advantage of RRR over
  PCA. 

  We integrated the relative decoding performance quantity over normalized log rank and correlated the resulting AUCs between datasets, matching observations by model, layer, and decoding
  task and retaining combinations available in all three datasets.

  Statistical significance was indicated at $p<0.05$. Cross-dataset AUC tests and Pearson correlation tests used unadjusted p-values; Holm correction was confined
  to the comparisons within each saturation panel.

\subsection{Significance Selection}

Neural targets were selected using held-out predictions from full-embedding ridge encoding models. For iEEG, we used five contiguous word folds and evaluated each electrode using its best-lag prediction correlation, defined as the sampled time point relative to word onset at which the correlation between predicted and observed neural activity was greatest. Null distributions were generated by circularly shifting responses within held-out folds. Best-lag (p)-values were corrected across electrodes using the Benjamini--Hochberg procedure. Once an electrode was selected, all sampled lags for that electrode were retained as separate RRR targets. Main analyses used the union of electrodes selected across configured model and layer significance runs.

For fMRI, voxel significance was assessed using predictions from models that held out complete training sessions, each consisting of unique stories and one repeated test story. The repeated test story was excluded from target selection. Null distributions were generated by circularly shifting responses within each story, followed by Benjamini--Hochberg correction. Significant voxels were retained as RRR targets.

\subsubsection{Results of Selection}

\begin{figure}[]
    \centering
    \includegraphics[width=\linewidth]{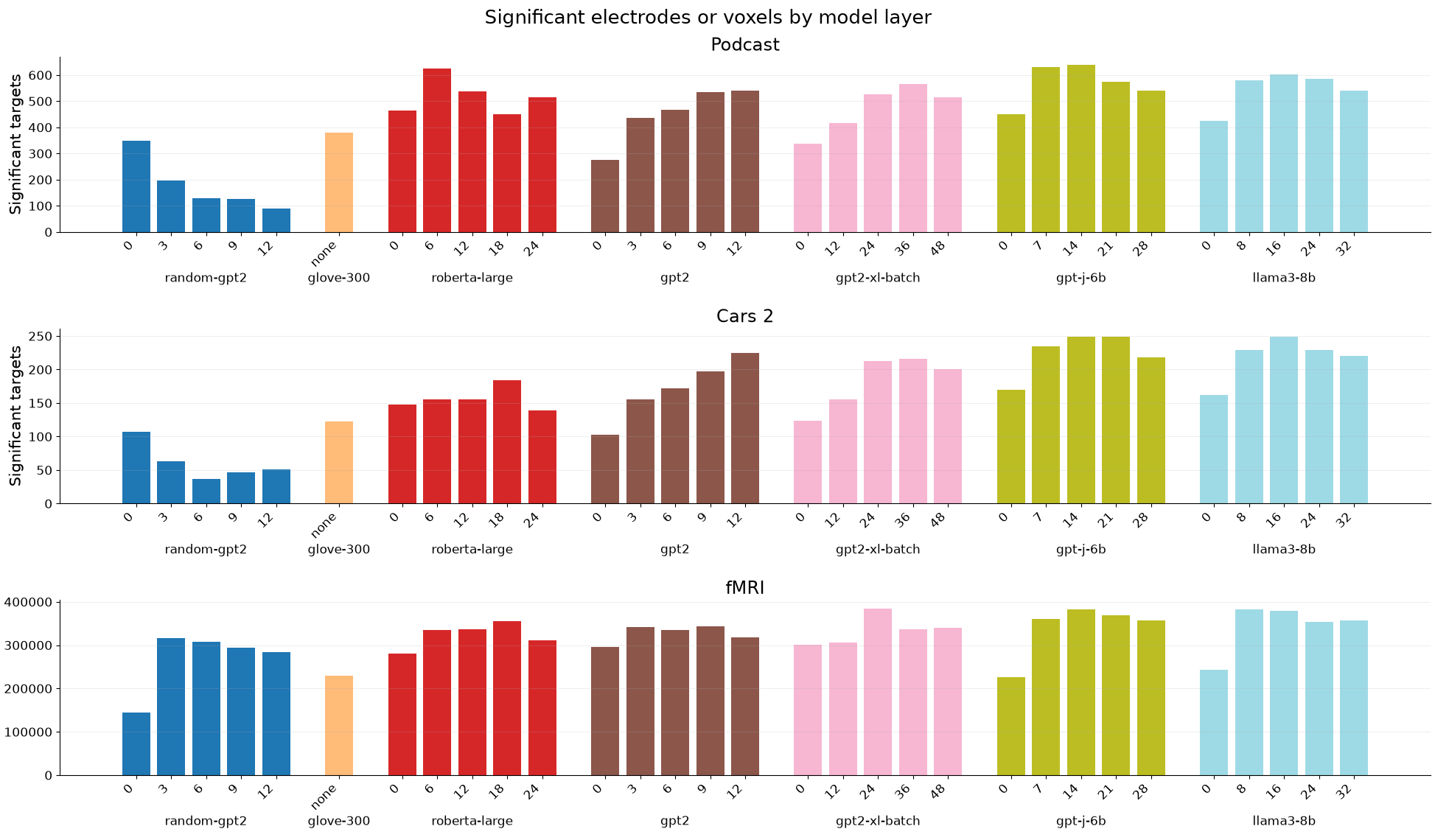}
    \caption{We use all targets deemed significant by at least one model–layer combination. See table \ref{tab:sigtargs} for full counts.}
    \label{fig:sigtargs}
\end{figure}

See figure \ref{fig:sigtargs} and table \ref{tab:sigtargs} for significance results.

\subsection{Encoding Model Performance}

\begin{figure}[t!]
    \centering
    \includegraphics[width=\linewidth]{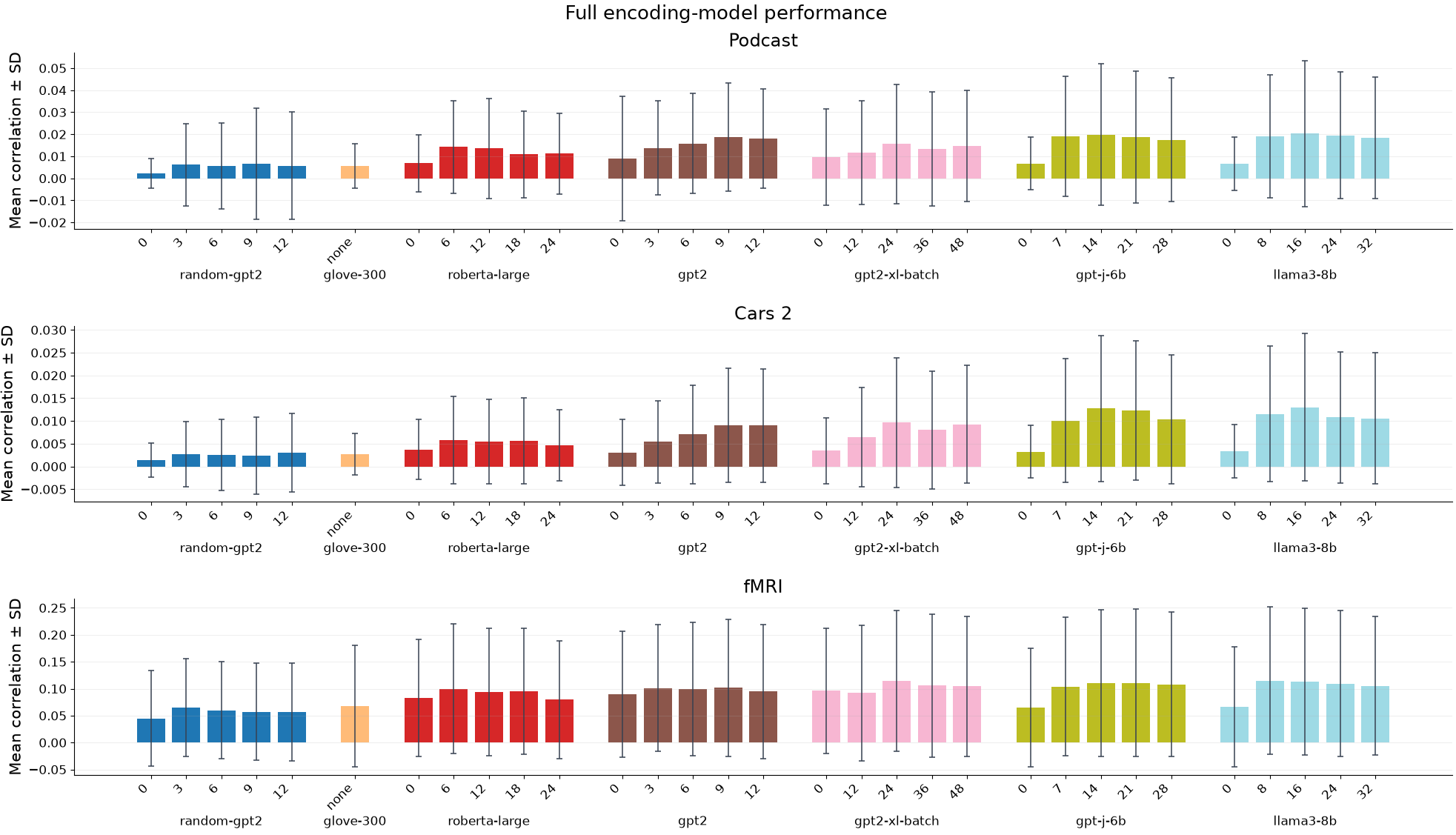}
    \caption{Mean encoding results of all of our model-layers across all of our datasets. We generally recover patterns typical of previous work where larger models outperform smaller models, and middle layers perform better than early or late layers. Performance is quite spread due to our liberal significance filtering.}
    \label{fig:encodingresults}
\end{figure}

See figure \ref{fig:encodingresults} for a summary figure.

\subsection{Linguistic Decoding Task Labels}
See table \ref{tab:language-decoding-features} for a complete list and description of tasks used.

\begin{table}[]

\caption{Number of significant targets (electrodes for Podcast and Cars 2, voxels for fMRI) for each model-layer. Union represents the final number of targets used for our analysis by taking all targets which any model-layer deemed significant.}
\label{tab:sigtargs}
\begin{tabular}{llrrrrr}
\toprule
Dataset & Model & 0\% & 25\% & 50\% & 75\% & 100\% \\
\midrule
Podcast & random-gpt2 & 350 & 197 & 128 & 127 & 90 \\
 & glove-300 & \multicolumn{5}{c}{381} \\
 & roberta-large & 463 & 626 & 538 & 450 & 515 \\
 & gpt2 & 275 & 435 & 467 & 535 & 541 \\
 & gpt2-xl-batch & 337 & 415 & 526 & 565 & 516 \\
 & gpt-j-6b & 450 & 630 & 638 & 574 & 540 \\
 & llama3-8b & 426 & 579 & 602 & 584 & 539 \\
 & Union & \multicolumn{5}{c}{1,200} \\
\midrule
Cars 2 & random-gpt2 & 107 & 63 & 37 & 47 & 51 \\
 & glove-300 & \multicolumn{5}{c}{122} \\
 & roberta-large & 148 & 156 & 155 & 184 & 139 \\
 & gpt2 & 103 & 156 & 172 & 197 & 225 \\
 & gpt2-xl-batch & 124 & 156 & 213 & 216 & 200 \\
 & gpt-j-6b & 170 & 235 & 249 & 249 & 218 \\
 & llama3-8b & 162 & 229 & 249 & 229 & 220 \\
 & Union & \multicolumn{5}{c}{439} \\
\midrule
fMRI & random-gpt2 & 144,979 & 316,842 & 307,353 & 294,695 & 284,196 \\
 & glove-300 & \multicolumn{5}{c}{229,249} \\
 & roberta-large & 280,635 & 335,494 & 336,063 & 356,103 & 311,036 \\
 & gpt2 & 296,089 & 342,502 & 334,932 & 343,213 & 317,423 \\
 & gpt2-xl-batch & 301,334 & 306,597 & 385,015 & 337,107 & 340,529 \\
 & gpt-j-6b & 226,388 & 360,222 & 382,378 & 368,703 & 357,002 \\
 & llama3-8b & 242,553 & 381,781 & 378,544 & 353,003 & 357,827 \\
 & Union & \multicolumn{5}{c}{618,587} \\
\bottomrule
\end{tabular}
\end{table}

\begin{table}[]
  \caption{Features decoded in the internal Cars 2, fMRI, and Podcast configurations.}
  \label{tab:language-decoding-features}
  \begin{center}
  \resizebox{\textwidth}{!}{%
  \begin{tabular}{llll}
  \bf FEATURE & \bf DATASET(S) & \bf MEASUREMENT AND CALCULATION & \bf CITATION(S) \\\hline
  Age of Acquisition
      & Podcast
      & Mean rated age at which the word is learned.
      & \citep{kupermanAgeofacquisitionRatings300002012} \\

  Arousal
      & fMRI, Podcast
      & Rated emotional activation associated with the word.
      & \citep{mohammadNRCVADLexicon2025} \\

  Association Entropy
      & Podcast
      & Entropy of the SWOW response-strength distribution for the cue word.
      & \citep{dedeyneSmallWorldWords2019,shannonMathematicalTheoryCommunication1948} \\

  Concreteness
      & Cars 2, fMRI, Podcast
      & Mean rating of how tangible or perceptually accessible the word's meaning is.
      & \citep{brysbaertConcretenessRatings402014} \\

  Content Word
      & Cars 2, fMRI, Podcast
      & Whether the word's universal part-of-speech category is a content-word category.
      & \citep{nivreUniversalDependenciesV22020, qiStanzaPythonNatural2020} \\

  Dependency Depth
      & Cars 2, fMRI, Podcast
      & Number of dependency-tree edges between the word and the sentence root.
      & \citep{nivreUniversalDependenciesV22020,qiStanzaPythonNatural2020} \\

  Dependency Length
      & Cars 2, fMRI, Podcast
      & Absolute linear distance in words between the word and its syntactic head.
      & \citep{gibsonLinguisticComplexityLocality1998,nivreUniversalDependenciesV22020,qiStanzaPythonNatural2020} \\

  Distance from Previous Occurrence
      & Cars 2, fMRI, Podcast
      & Number of tokens since the preceding occurrence of the normalized word.
      & Derived from aligned transcript \\

  Volume
      & Cars 2, fMRI, Podcast
      & Root-mean-square audio amplitude in a fixed window following word onset.
      & \citep{wangBrainTreebankLargescale2024} \\

  Named Entity
      & Podcast
      & Whether the word belongs to a named-entity span.
      & \citep{qiStanzaPythonNatural2020} \\

  Intonation Unit End
      & fMRI, Podcast
      & Whether the word is the final word of an automatically detected intonation unit.
      & \citep{rollPSSTProsodicSpeech2023} \\

  Intonation Unit Length
      & Podcast
      & Number of words in the word's automatically detected intonation unit.
      & \citep{rollPSSTProsodicSpeech2023} \\

  Intonation Unit Onset
      & fMRI, Podcast
      & Whether the word is the first word of an automatically detected intonation unit.
      & \citep{rollPSSTProsodicSpeech2023} \\

  Intonation Unit Position
      & fMRI, Podcast
      & Normalized position of the word within its automatically detected intonation unit.
      & \citep{rollPSSTProsodicSpeech2023} \\

  Pitch
      & Cars 2
      & Mean fundamental-frequency estimate in the audio window associated with the word.
      & \citep{wangBrainTreebankLargescale2024} \\

  Part of Speech
      & Cars 2, fMRI, Podcast
      & Universal grammatical category assigned to the word.
      & \citep{nivreUniversalDependenciesV22020,qiStanzaPythonNatural2020} \\

  Post-Word Pause
      & Cars 2, fMRI, Podcast
      & Time between the word's offset and the following word's onset.
      & Derived from aligned timestamps \\

  Pre-Word Pause
      & Cars 2, fMRI, Podcast
      & Time between the preceding word's offset and the current word's onset.
      & Derived from aligned timestamps \\

  Prior Word Count
      & Cars 2, fMRI, Podcast
      & Number of earlier occurrences of the normalized word in the stimulus.
      & Derived from aligned transcript \\

  Semantic Diversity
      & Podcast
      & Variability of the linguistic contexts in which the word occurs.
      & \citep{hoffmanSemanticDiversityMeasure2013} \\

  Sentence End
      & Cars 2, fMRI, Podcast
      & Whether the word is the final word of its sentence.
      & \citep{nivreUniversalDependenciesV22020,qiStanzaPythonNatural2020} \\

  Sentence Onset
      & Cars 2, fMRI, Podcast
      & Whether the word is the first word of its sentence.
      & \citep{nivreUniversalDependenciesV22020,qiStanzaPythonNatural2020} \\

  Sentence Position
      & Cars 2, fMRI, Podcast
      & Normalized position of the word within its sentence.
      & \citep{nivreUniversalDependenciesV22020,qiStanzaPythonNatural2020} \\

  Sense Count
      & Podcast
      & Number of WordNet synsets associated with the word's lemma and part of speech.
      & \citep{princeton2010wordnet} \\

  Speaker Identity
      & Cars 2, Podcast
      & Native speaker annotation for Cars 2 and predicted diarization cluster for Podcast.
      & \citep{wangBrainTreebankLargescale2024,bredinPyannoteaudioNeuralBuilding2019} \\

  Sensorimotor Strength
      & Podcast
      & Maximum rated strength across the Lancaster perceptual and action dimensions.
      & \citep{lynottLancasterSensorimotorNorms2020} \\

  Entropy
      & Cars 2, fMRI, Podcast
      & Entropy of GPT-2's next-token distribution before observing the word.
      & \citep{shannonMathematicalTheoryCommunication1948,radford2019language} \\

  Surprisal
      & Cars 2, fMRI, Podcast
      & Negative log probability assigned by GPT-2 to the observed word given its preceding context.
      & \citep{radford2019language} \\

  Valence
      & fMRI, Podcast
      & Rated positive or negative emotional association of the word.
      & \citep{mohammadNRCVADLexicon2025} \\

  Word Duration
      & Cars 2, fMRI, Podcast
      & Difference between the aligned word offset and onset times.
      & Derived from aligned timestamps \\

  Word in Previous 5
      & Cars 2, fMRI, Podcast
      & Whether the normalized word occurs among the five preceding tokens.
      & Derived from aligned transcript \\

  Word Length
      & Cars 2, fMRI, Podcast
      & Number of orthographic characters in the normalized word.
      & Derived from aligned transcript \\

  Word Prevalence
      & Podcast
      & Transformed proportion of participants who reported knowing the word.
      & \citep{brysbaertWordPrevalenceNorms2019} \\

  Word Zipf Frequency
      & Cars 2, fMRI, Podcast
      & Logarithmic frequency per billion words obtained from \texttt{wordfreq}.
      & \citep{speer2022wordfreq} \\
  \end{tabular}%
  }
  \end{center}
  \end{table}

\end{document}